\documentclass{article} % For LaTeX2e
\usepackage{iclr2023_conference,times}

\usepackage{amsmath,amsfonts,bm}

\def\eqref#1{equation~\ref{#1}}
\def\1{\bm{1}}

\DeclareMathAlphabet{\mathsfit}{\encodingdefault}{\sfdefault}{m}{sl}
\SetMathAlphabet{\mathsfit}{bold}{\encodingdefault}{\sfdefault}{bx}{n}

\usepackage{hyperref}
\usepackage{url}

\usepackage[utf8]{inputenc} % allow utf-8 input
\usepackage[T1]{fontenc}    % use 8-bit T1 fonts
\usepackage{hyperref}       % hyperlinks
\usepackage{url}            % simple URL typesetting
\usepackage{booktabs}       % professional-quality tables
\usepackage{amsfonts}       % blackboard math symbols
\usepackage{nicefrac}       % compact symbols for 1/2, etc.
\usepackage{microtype}      % microtypography
\usepackage{xcolor}         % colors

\usepackage{graphicx}

\usepackage{tikz}
\usepackage{comment}
\usepackage{amsmath,amssymb}

\usepackage{booktabs}
\usepackage{multirow}
\usepackage{multicol}
\usepackage{dsfont}
\usepackage{colortbl}
\usepackage{lipsum}
\usepackage{xspace}
\usepackage{makecell}
\usepackage{color}
\usepackage{enumitem}
\usepackage{amsthm}
\usepackage{wasysym}
\setitemize{noitemsep,topsep=0pt,parsep=0pt,partopsep=0pt}

\usepackage{float}

\newtheorem{theorem}{Theorem} % [section]
\newcommand{\vpara}[1]{\vspace{0.07in}\noindent\textbf{#1}\xspace}

\definecolor{mygray}{gray}{.9}
\definecolor{mypink}{rgb}{.99,.91,.95}
\definecolor{mygreen}{rgb}{.52,.73,.30}
\definecolor{myblue}{rgb}{.39,.58,.93}
\definecolor{mycyan}{cmyk}{.3,0,0,0}
\definecolor{mylakeblue}{rgb}{.0,.749,1.}
\definecolor{mypurple}{rgb}{.729,.333,.827}
\definecolor{myclassicblue2}{rgb}{.117,.565,1.}
\definecolor{myclassicblue}{rgb}{1.0,.3176,.3176}

\title{Self-Supervised Category-Level Articulated Object Pose Estimation with Part-Level SE(3) Equivariance}

\author{
Xueyi Liu$^{1,7}$,
Ji Zhang$^{2}$,
Ruizhen Hu$^{3}$,
Haibin Huang$^{4}$,
He Wang$^{5}$,
Li Yi$^{1,6,7}$
\\
$^{1}$Tsinghua University~~
$^{2}$Fudan University~~
$^{3}$Shenzhen University~~
$^{4}$Kuaishou Technology~~
\\
$^{5}$Peking University~~
$^{6}$Shanghai Artificial Intelligence Laboratory~~
$^{7}$Shanghai Qi Zhi Institute
}

\iclrfinalcopy % Uncomment for camera-ready version, but NOT for submission.
\begin{document}

\maketitle

% !TEX root = ../main.tex

\begin{abstract}
  Category-level articulated object pose estimation aims to estimate a hierarchy of articulation-aware object poses
%   per-part pose, joint parameter and joint states 
    of an unseen articulated object 
  from a known category.
%   achieve 
%   the factoraization of each part shape, 
%   achieve 
%   a hierarchical 
%   understanding of ''
%   pose and shape disentanglement from object-level to part-level for an unseen articulated object from the known category.
%   both part-level and object-level   for an unseen articulated object from the known category.
%   understanding of the input shape from both part-level and object-level  for an unseen articulated object from the known category.
%   the input shape from both part-level such as per-part 6D pose and object-level such as joint parameters for an unseen articulated object from the known category.
%   with both part-level properties such as per-part 6D pose and object-level properties such as joint parameters decomposed. 
%   part pose estimation aims to predict 6D part pose for each part of an unseen articulated object from the known category.
%   of an unseen articulated object from the known category. 
  To reduce the heavy annotations needed for supervised learning methods, we present a novel self-supervised strategy that solves this problem without any human labels. 
Our key idea is to factorize canonical shapes and articulated object poses from input articulated shapes through part-level equivariant shape analysis.
% euqivariant shape analysis.
% \textcolor{myblue}{
% Considering the complex structure of articulated objects, we introduce the concept of part-level SE(3) equivariance to make the problem solvable in a self-supervised fashion. The factorization is achieved by 1) using invariant features to encourage both global and local category-level canonical spaces to emerge automatically; 2) using part-level equivariant features to isolate the effect of one part on the other to reduce the difficulty of pose decomposition. 
% }
Specifically, we first introduce the concept of part-level SE(3) equivariance and devise a network to learn features of such property. 
% Then, by leveraging equivariant features for the prediction of pose-related properties and by encouraging invariant canonical shapes to emerge from invariant features, the difficulty of category-level articulated pose estimation is reduced so that it is solvable in a self-supervised manner. 
Then, through a carefully designed fine-grained pose-shape disentanglement strategy, we expect that canonical spaces to support pose estimation could be induced automatically.
Thus, we could further predict articulated object poses as per-part rigid transformations describing how parts transform from their canonical part spaces to the camera space. 
% how parts are assembled together 
% the object assembling process from their canonical spaces to the camera space. 
% for defining poses 
% and further the category-level articulated object poses could be induced 
% by the network 
% automatically. 
Extensive experiments demonstrate the effectiveness of our method on both complete and partial point clouds from 
% popular 
synthetic and real articulated object datasets.
The project page with code and more information can be found at: 
% and visualizations: 
\href{https://equi-articulated-pose.github.io}{equi-articulated-pose.github.io}.
\end{abstract}

\section{Introduction}
% part pose estimation
% 
% the demand for a self-supervised method for part pose estimation
% challenges of the task
% the demand for a self...

% \eric{Our key contribution is to present a self-supervised framework achieving part shape, part structure, joint, and part pose disentanglement. We achieve this leveraging part-level SE(3) equivariance from a novel equivariant network design.}

% Part pose estimation for articulated objects 
% \summary{Background.} 
Articulated object pose estimation
is a crucial and fundamental computer vision problem with a wide range of applications in robotics, human-object interaction, and augmented reality~\cite{katz2008manipulating,mu2021maniskill,labbe2021single,jiang2022ditto,goyal2022human,li2020detailed}. 
% Interacting with such articulated objects with the understanding of part-level poses and 
% The requirement for understanding an articulated objects by parts with inferred part-level poses is beyond pose estimation for rigid objects~\cite{tremblay2018deep,xiang2017posecnn,sundermeyer2018implicit,wang2019normalized}.
% Moreover, we aim at a hierarchical understanding for articulated obje
Different from 6D pose estimation for rigid objects~\cite{tremblay2018deep,xiang2017posecnn,sundermeyer2018implicit,wang2019normalized}, articulated object pose estimation requires a hierarchical pose understanding on both the object-level and part-level~\cite{li2020category}. 
% a hierarchical pose understanding on both the object-level and part-level~\cite{li2020category}.
% Existing articulated object pose estimation works can be mainly cast into two lines: instance-level pose estimation with the assumption of available object CAD models and joint parameters~\cite{katz2008manipulating,katz2013interactive,hausman2015active}; category-level pose estimation which can generalize to unseen objects from a known category~\cite{li2020category}. 
% It is beyond pose estimation for an rigid object~\cite{tremblay2018deep,xiang2017posecnn,sundermeyer2018implicit,wang2019normalized}, and  estimation for each individual part, as well.
% There are a lot of works trying to do part pose estimation whether in the instance level with the assumption of available object CAD models and joint parameters~\cite{katz2008manipulating,katz2013interactive,hausman2015active}, or in the category-level by defining a category-level reference frame for articulated objects that could generalize to unseen objects from the same category~\cite{li2020category}.
% Category-level pose estimation assumes no a
This problem has been long studied on the instance level where an exact CAD model is required to understand the pose of a specific instance. Recently, there is a trend in estimating category-level object pose such that the algorithm can generalize to novel instances. 
Despite such merits, supervised category-level approaches always assume rich annotations that are extremely expensive to acquire~\cite{li2020category,chi2021garmentnets,9670684}. 
% are extremely expensive due to the required heavy annotation efforts. 
% Therefore, 
% we aim to tackle this problem under a self-supervised setting instead. 
To get rid of such restrictions, we tackle this problem under a self-supervised setting instead. 
% need a large amount of training data to achieve intra-category generalization. 
% Thus, those methods 
% Among them, category-level pose estimation has wider application scenarios compared with instance-level ones considering new objects that a vision system may encounter every day. 
% However, a category-level approach also indicates a larger amount of training data to achieve intra-category generalization. 
% That's what make a supervised category-level method extremely expensive due to the required heavy annotation efforts. %  in the training process. 
% \zz{need to specify the difference between category-level and instance-level in intro?}

% \summary{Problem Definition.} 
Given a collection of unsegmented articulated objects in various articulation states with different object poses,  
our goal is to design a network that can acquire a category-level articulated object pose understanding in a self-supervised manner without any human labels such as pose annotations, segmentation labels, or
% category-level 
reference frames for pose definition. 
The self-supervised category-level articulated object pose estimation problem is highly ill-posed since it requires the knowledge of object structure and per-part poses, which are usually entangled with part shapes.
% Various entangled information in the articulated objects makes the self-supervised decomposition a highly ill-posed problem for neural networks. 
% It is a quite challenging problem considering the complexity of articulated objects and the unsupervised setting. 
Very few previous works try to solve such a problem or even similar ones. % ~\cite{li2021leveraging,lica}. 
The most related attempt is the work of~\cite{li2021leveraging}.
It tackles the unsupervised category-level pose estimation problem but just for rigid objects.
% using a shape reconstruction-based self-supervised task. 
% The key idea is to 
They leverages SE(3) equivariant shape analysis to disentangle the global object pose and shape information
% use SE(3) equivariant features 
so that a category-aligned canonical object space can emerge. 
% It further supports the category-level object pose prediction. 
This way the category-level object poses could be automatically learned by predicting a transformation from the canonical space to the camera space. 
Going beyond rigid objects, estimating articulated object poses demands more than just global pose and shape disentanglement. 
It requires 
% a hierarchical 
% % pose understanding 
% factorization
a more fine-grained disentanglement of part shape, object structure such as part adjacency relationship, joint states, part poses, and so on. 
To achieve such fine-grained disentanglement, we propose to leverage part-level SE(3) equivariant shape analysis. Especially, we introduce the concept of part-level SE(3) equivariant features to equip equivariance with a spatial support.
The part-level SE(3) equivariant feature of a local region should only change as its parent part transforms but should not be influenced by the transformation of other parts. This is in contrast to the object-level  SE(3) equivariant feature for a local region, which is influenced by both the region's parent part and other parts. To densely extract part-level SE(3) equivariant features from an articulated shape, we propose a novel pose-aware equivariant point convolution operator.
Based on such features, 
% to define and predict articulated object poses, 
we are able to achieve a fine-grained disentanglement which learns three types of information from input shapes:
1) \emph{Canonical part shapes}, which are invariant to input pose or articulation changes and are category-aligned to provide a consistent reference frame for part poses;
% including the  canonical object space for recovering object structure and orientation and a set of  canonical part spaces for determining 
% part shapes and poses; %both of which are invariant to object poses;
%part space for each part with part poses canonicalized and the canonical object space with part associations and joint parameters recovered, the object orientation and joint states canonicalized~\cite{li2020category}. 
% Both of them should be invariant to part rigid transformations.
%We expect both of them invariant to object poses.
% They can further sever as reference frames for part pose definition. 
% category-level canonical part spaces and the canonical object space,
% recover part connection and kinematic relationship, as well as per-part rigid transformations across them. 
2) \emph{Object structure}, which is also invariant to input pose or articulation changes and contains structural information about the
% including the kinematic chain of the articulated object which encodes 
part adjacency relationships, part transformation order, and joint parameters such as pivot points;
% and transformation order in the canonical object space, and joint parameters such as pivot points;
% and joint parameters and the kinematic chain that recovers the part transformation order in the canonical object space. 
% Such information 
% % encoding 
% % inter-part relationship and kinematic constraints 
% allows us to predict part poses in an articulation-aware manner. 
%They describe the pose-invariant structure of the object in the canonical object space. 
% 
3) \emph{Articulated object pose}, which is composed 
% by 
of
a series of estimated transformations. Such transformations include 
per-part rigid transformations which assembles canonical part shapes into a canonical object shape, per-part articulated transformation which articulates the canonical object shape to match the input articulation state,
% a from canonical part spaces to the canonical object space, 
% per-part articulated transformations in the canonical object space, 
and a base part rigid transformation transforming the articulated canonical object to the camera space. 
% Each type of information is learned by networks from equivariant or invariant features depending on their equivariant/invariant natures. 
%%%%%%%%%%%
% The disentangled information enables us to recover the observed shape's generation in a part-by-part manner. 
% the generation of the observed shape part-by-part. 
% }
% Therefore, we can further adopt a self-supervised shape reconstruction task to guide the network learning. 
%%%%%%%%%%%%
% We further adopt a self-supervised part-by-part shape reconstruction task to guide the network learning by combining the above disentangled information to recover input articulated shapes.
To allow such disentanglement, we guide the network learning through a self-supervised part-by-part shape reconstruction task that combines the disentangled information to recover the input shapes.
With the above self-supervised disentanglement strategy, our method demonstrates the possibility of estimating articulated object poses in a self-supervised way for the first time. 
Extensive experiments prove its effectiveness on both complete point clouds and partial point clouds from various categories covering both synthetic and real datasets. 
On the Part-Mobility Dataset~\cite{wang2019shape2motion}, our method without the need for any human annotations can already outperform the iterative pose estimation strategy with ground-truth segmentation masks on both complete and partial settings by a large margin, \emph{e.g.} reduce the rotation estimation error by around 30 degrees on complete shapes and by 40 degrees on partial shapes. 
% around  30 degrees of rotation estimation error less than ICP on complete shapes and around less 40 degrees on partial shapes. 
% reduce the average rotation angle error from 46.11$^\circ$ to 7.90$^\circ$ on the complete shapes, and from 48.55$^\circ$ to 8.36$^\circ$ on partial data, closing the gap between supervised approaches to around 3$\sim$5 degrees. 
% \eric{update with the average number}
Besides, our method can perform on par with to or even better than supervised methods like NPCS~\cite{li2020category}.
% For instance, we can achieve an average mean rotation angle error of 7.90$^{\circ}$ compared with 5.38$^{\circ}$  of NPCS on complete point clouds. 
% With respect to the average mean rotation angle error, our result on both complete shapes and partial shapes are comparable to those of NPCS
For instance, we can achieve an average of 7.9$^{\circ}$ rotation estimation error on complete shapes, comparable to NPCS's 5.8$^{\circ}$ error.  
% can reduce the rotation estimation error to around 7.9 degrees on complete shapes, comparable to NPCS's 5.8 degree error. 
We can even outperform NPCS on some specific categories such as partial Eyeglasses. 
% \eric{update with the average number, don't need to be too detailed.}
% we can even sometimes outperform \textbf{supervised} methods. 
% our method can sometimes achieve better results than \textbf{supervised} baselines. 
% On the Safe-real category~\cite{liu2022hoi4d}, we can arrive at 4.4$^\circ$ joint direction prediction error, more accurate than 14.2$^\circ$ prediction error of the supervised NPCS. 
% For instance, we can predict more accurate joint directions 
% we can achieve more accurate joint direction prediction results 
% with an average 4.4$^\circ$ error on the Safe category compared with that of supervised NPCS (14.2$^\circ$). 
% Moreover, we further demonstrate the robustness of our strategy to object occlusions by testing it on \textbf{partial} point clouds. 
% We can obtain better results 
% We can achieve better performance 
% than supervised NPCS more 
% It could perform better than the supervised baseline more 
% frequently on partial data. 
% e.g. more accurate part pose prediction on the Eyeglasses category. 
% on \textbf{partial} point clouds, 
% Our hierarchical pose understanding can also naaturally avoid the part symmetry problem which would results in pose ambiguity and leads to the performance degeneration of traditional unsupervised methods.
% even comparable to supervised methods. 
Finally, we prove the effectiveness of our part-level SE(3) equivariance design and the fine-grained disentanglement strategy
% and the hierarchical pose understanding
in the ablation study. 
% In the ablation study, we further prove the effectiveness of the part-level SE(3) equivariant network design as well as the value of part kinematics modeling. 
Our main contributions are summarized as follows: 
\begin{itemize}[leftmargin=.5cm]
    \item To our best knowledge, we are the first that tackles the self-supervised articulated object pose estimation problem. 
    % trying to design a solution for articulated object pose estimation in a self-supervised way. %Our method can even achieve results comparable to supervised methods sometimes. 
    % trying to solve the unsupervised articulated object pose estimation problem with a strategy proposed achieving results even comparable to supervised methods.  x
    \item We design a pose-aware equivariant point convolution operator to 
    % feature communication strategy to 
    % extract 
    learn part-level SE(3)-equivariant features.
    % acquire part-level SE(3)-equivariance based on a popular SE(3)-equivariant network.
    \item We propose a self-supervised framework to achieve the disentanglement of canonical shape, object structure, and articulated object poses. 
    % part shape, object structure, and articulated object pose. %  based on a part-by-part reconstruction task. 
\end{itemize}

\section{Related Works}

% \textcolor{orange}{Difference from previous works...}

% \begin{itemize}
%     \item Unsupervised pose-aware part decomposition for 3D articulated objects. (\textcolor{orange}{joint, category-common pose prior, joint/pivot point for each part/each pair of parts, part proposal, adversarial loss for shape reconstruction, disentangle part pose from part shape, shape representation --- implicit field or point clouds})
    
%     % unsupervised part decomposition 
    
%     It seems that this work $\approx$ BSP-Net + Pose-aware (part decomposition)
%     \item BSP-Net, BAE-Net (\textcolor{orange}{part proposal, alignment assumption for input shapes})
%     \item Equivariant Point Cloud Analysis via Learning Orientations for Message Passing (\textcolor{orange}{orientation-based part-level equiv? --- inv to orientation + orientation $\rightarrow$ equiv to orientation})
% \end{itemize}

\vpara{Unsupervised Part Decomposition for 3D Objects.}
Decomposing an observed 3D object shape into parts in an unsupervised manner is a recent interest in shape representation learning. 
% Previous works mainly adopt a generative strategy to design a shape reconstruction-based self-supervised learning task. 
% The previous main stream 
Previous works
always tend to adopt a generative shape reconstruction task to self-supervise the shape decomposition. 
% Those methods 
They often choose to represent parts via learnable primitive shapes~\cite{tulsiani2017learning,kawana2020neural,yang2021unsupervised,paschalidou2021neural,deng2020cvxnet,zhu2020adacoseg,chen2020bsp}
% , where the representative ability of part shape decoder is either restricted to certain primitive types like cuboids~\cite{yang2021unsupervised}, start domains~\cite{kawana2020neural}, and superquadrics~\cite{paschalidou2019superquadrics}, 
or 
% not restricted like the 
non-primitive-based implicit field representation~\cite{chen2019bae,kawana2022uppd}. 
% Moreover, 
% For shapes from the same category, semantic-awareness and cross-instance consistency are 
% % good properties 
% widely pursued for decomposed parts~\cite{chen2019bae,chen2020bsp,zhu2020adacoseg}.
% the decomposed parts are also wished to be semantic-aware, thus consistent across different shapes~\cite{chen2019bae,chen2020bsp,zhu2020adacoseg}. 
% For -level shape co-segmentation, the decomposed parts are also wished to be semantic-aware, thus consistent for shapes in a same category. 
% Therefore,
Shape alignment is a common assumption of such methods to 
achieve consistent decomposition across different shapes.

\vpara{Articulated Object Pose Estimation.} 
Pose estimation for articulated objects aims to acquire a fine-grained understanding of target articulated objects from both the object level and the part level. 
The prior work~\cite{li2020category} proposes to estimate object orientations, joint parameters, and per-part poses in a fully-supervised setting. 
They define Articulation-aware Normalized Coordinate Space Hierarchy (ANCSH), composed of the canonical object space and a set of canonical part spaces, 
as a consistent %  canonical 
representation for articulated objects to support pose estimation.
% for defining poses. 
% ANCSH is further composed of two-level hierarchy of of canonical spaces, including the Normalized Articulated Object Coordinate Space (NAOCS) and a set of Normalized Part Coordinate Spaces (NPCS) as the object level canonical space and part-level canonical spaces respectively. 
In this work, we also want to estimate a hierarchy of articulation-aware object poses but in a totally unsupervised setting. 
% we want to estimate a hierarchy of articulation-aware object poses but in a totally unsupervised setting. 
% We also define poses on two levels of canonical spaces but propose to let them  automatically induced by the network during the learning. 
Instead of hand-crafting normalized coordinate spaces, we wish to let them be automatically induced during learning. 
% Such vision could be accomplished thanks to the part-level SE(3) invariant features that can be learned sby our part-level SE(3) equivariant network. 

\vpara{SE(3) Equivariant Networks.}
Recently, there is a trend of pursuing SE(3)-equivariant and invariant features through network design~\cite{weiler20183d,thomas2018tensor,fuchs2020se,zhao2020quaternion,chen2021equivariant}.
% SE(3)-equivariant and invariant features are more and more pursed by recent networks~\cite{weiler20183d,thomas2018tensor,fuchs2020se,zhao2020quaternion,chen2021equivariant}.
Equivariance 
% or approximate equivariance 
is achieved by designing kernels~\cite{thomas2018tensor,fuchs2020se} or designing feature convolution strategies~\cite{chen2021equivariant,zhao2020quaternion}.
% specially designed kernels such as spherical harmonics~\cite{thomas2018tensor,fuchs2020se} or through other strategies~\cite{chen2021equivariant,zhao2020quaternion}.
%%%%%%%%%%
% For rigid objects, features extracted by those equivariant networks will change equivalently to the SE(3) transformation of the input shape. 
%%%%%%%%%%
% equivalent to the pose change of the input.
% Leveraging such equivariant features, a recent work proposes to estimate category-level 6D pose for rigid 
% % input 
% objects~\cite{li2021leveraging} in an unsupervised manner.
% based on a shape reconstruction task.
% based on a shape reconstruction task. 
% Using the invariant feature, they can enforce the emergence of a category-level reference frame for shapes in the same category.
In this work, we design our part-level SE(3) equivariant feature network based on Equivariant Point Network~\cite{chen2021equivariant} for articulated object pose estimation.
% to better suspport part-based properties estimation. 
% We highlight that part-level SE(3) equivariant features of a specific part would only change equivalently according to its own rigid transformations, while remains invariant to other parts'. 
% Such properties cannot be  
Common SE(3) equivariant feature of a local region would be affected by both its parent part's and other parts' rigid transformations. 
By contrast, its part-level SE(3) equivariant feature would only
be affected by its parent part. 
% change 
% equivalently according to the parent part. 
% but would not be affected by other parts. 
% common SE(3) equivariant features are only equivariant to the whole object's transformations. 
% In this work, we target at articulated objects and want to estimate part poses from the input shape. 
% To seek for part-level equivariant features, we propose a strategy to improve the adopted SE(3)-equivariant network. 
% Then based on equivariant features, we design our motion-based part-by-part reconstruction task to self-supervise the network's learning process. 
% on the adopted SE(3)-equivariant network 
% That indicates 
% To estimate equivariant part pose, we wish per-point equivariant features to have such equivariant property, which could not be achieved by directly using equivariant networks to process input shapes. 
% Thus, a pose-aware inter-part equivariant feature communication strategy is proposed to achieve this goal. 
% In this work, we target at articulated objects, whose parts 
% move from rigid objects to articulated objects, indicating a task composition of part decomposition, part pose estimation and shape reconstruction.
% Moreover, we propose a pose-aware inter-part feature communication strategy given the estimated part pose as input to get better SE(3)-equivariant features with contextual articulation state change disentangled.

% !TEX root = ../main.tex

\section{Method}

% outline for the proposed method
% \begin{itemize}
%     \item Part proposal module.
%     \item Part motion prediction strategy.
%     \begin{itemize}
%         \item Joint modeling strategy (point-based strategy \& assumptions and limitations)
%     \end{itemize}
%     \item Part-level equivariance. 
%     \begin{itemize}
%         \item Method.
%         \item Motivation.
%     \end{itemize}
% \end{itemize}

\begin{figure*}[ht]
    \centering
      \includegraphics[width=0.90\textwidth]{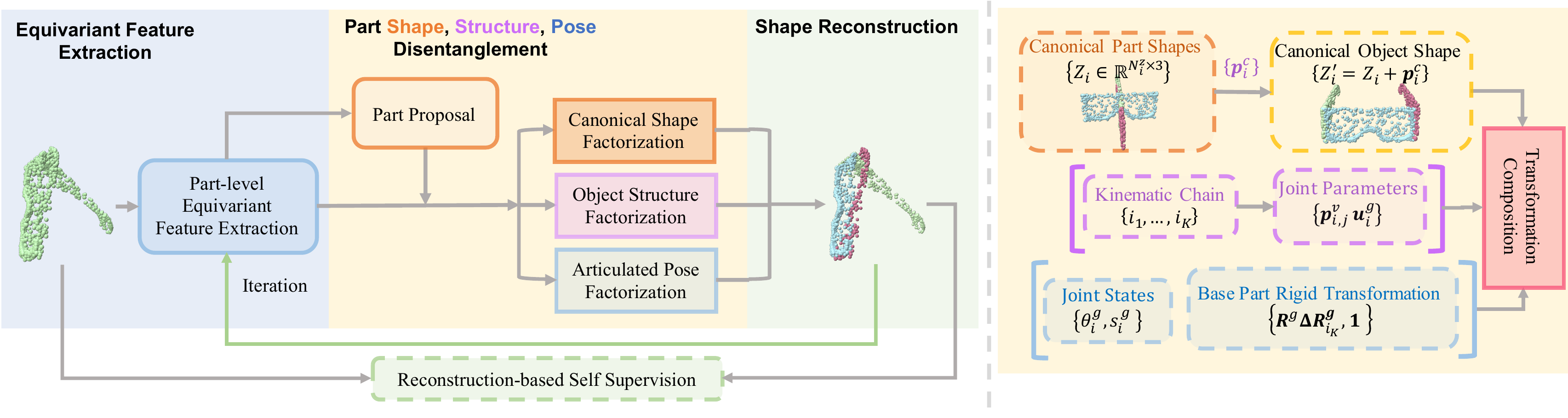} % overall-pipeline-z
      \vspace{-10pt}
    \caption{
      \footnotesize
    %   \textcolor{}{}
    % \textcolor{myblue}{
      Overview of the proposed self-supervised articulated object pose estimation strategy.
    The method takes a complete or partial point cloud of an articulated object as input, factorizes canonical shapes, object structure, and the articulated object pose from it. 
    The network is trained by a shape reconstruction task. 
    % Part-level SE(3) equivariant features are learned  by iterating between part pose estimation and pose-aware equivariant point convolution. 
    \textbf{Left:} A high-level abstraction of our pipeline. \textbf{Right:} An illustrate of decomposed information for shape reconstruction. 
      \textcolor{mygreen}{Green} lines (\textcolor{mygreen}{$\leftarrow$}) denote the iterative pose estimation process. 
    }
    \label{fig_overall_pipeline}
    % \vspace{-5pt}
    \vspace{-10pt}
\end{figure*}

% Our key contribution is to present a self-supervised framework achieving part shape, part structure, joint, and part pose disentanglement. We achieve this leveraging part-level SE(3) equivariance from a novel equivariant network design.

% \zz{Not necessary to use a separate section for Overview?}
% 
We present our method for self-supervised category-level articulated object pose estimation. 
We first propose to learn part-level SE(3) equivariant features through a novel pose-aware equivariant point convolution module (sec.~\ref{sec_method_revisit}). 
Based on such features, we then design a disentanglement strategy to factorize an arbitrarily posed 3D point cloud into three types of information. Such information includes a canonical shape with category-aligned pose and articulation state, 
% canonical spaces, 
the object structure describing the
part adjacency and joints, as well as the articulated object pose (sec.~\ref{sec_method_ssl_method}).
We find part-level SE(3) equivariant features are key to achieve the factorization above.
Further, we adopt a part-by-part shape reconstruction task that combines the factorized information for 
% input 
shape reconstruction to self-supervise the factorization (sec.~\ref{sec_method_self_supervised_task}). 
Our method assumes a category-level setting where input shapes have the same kinematic chain. 
For notations frequently used in the following text, $N$, $C$, and $K$ denote the number of points, feature dimension, and the number of parts per shape respectively. 
\subsection{Part-level SE(3)-equivariant Network}
\label{sec_method_revisit}
% \todo{need to say the pose : from the canonical frame to observed camera frame?}
% \todo{parts, points, rigid transformations}
% \summary{Introduction}
% Part-level SE(3)-equivariant point cloud network $\phi(\cdot, \cdot)$ operates on a point cloud with segmentation labels $X = \{ X_i\vert 1\le i \le K\}$ and part pose $P = \{ P_i \vert 1\le i\le K\}$ for per-part part-level SE(3)-equivariant feature 
% $F = \{ F_i \in \mathcal{F} \vert 1\le i\le K \}$, where $\mathcal{F}$ is an arbitrary feature domain, $K$ is the number of parts. \subsummary{Definition}
% Part-level equivariant feature $F_i$ of part $i$
% Part-level equivariant feature $F_i$ of part $i$ would change equivalently to the rigid transformation part $i$, but invariant to those of other parts. 
% Part-level equivariant feature $F_i$ would change equivalently to the rigid transformation of part $i$, but invariant to that of others. 
% Part-level SE(3)-equivariant point cloud network $\phi(\cdot)$ operates on a segmented point cloud $X$ with per-point pose $P$ for per-part part-level SE(3)-equivariant feature $F = \{ F_i \in \mathcal{F} \vert 1\le i\le K \}$, where $\mathcal{F}$ is an arbitrary feature domain.  \subsummary{Definition}
% \textcolor{myblue}{
% elaborate our 
We first elaborate on our part-level SE(3) equivariant network.
% the designed network for per-point part-level SE(3) equivariant feature extraction. 
The network $\phi(\cdot)$ operates on a point cloud 
% $X = \{X_i\vert 1\le i\le N \}$ 
% with 
$X =  \{ \mathbf{x}_i \vert 1\le i\le N \}$ 
with per-point pose and outputs part-level SE(3) equivariant features for all points 
$F = \{F_i=\phi(X)[i] \vert 1\le i\le N \}$. Here the pose of a point refer to the pose of that point's parent part.
%, where $\mathcal{S}$ is the feature domain. 
% $F = \phi(X)[i] \in \mathcal{F} \vert 1\le i\le N \}$, 
% where $\mathcal{F}$ is an arbitrary feature domain.
% $N$ is the number of points in the point cloud. 
% Part-level SE(3)-equivariant point cloud network $\phi(\cdot)$ operates on a segmented point cloud $X = \{ \mathbf{x}_i \vert 1\le i\le N \}$ with per-point pose $P$ for per-point part-level SE(3)-equivariant feature $F = \{ F_i = \phi(X)[i] \in \mathcal{F} \vert 1\le i\le N \}$, where $\mathcal{F}$ is an arbitrary feature domain, $N$ is the number of points in the point cloud. 
% \subsummary{Definition}
We introduce the concept of part-level equivariant feature to differentiate from object-level equivariant features in~\cite{chen2021equivariant}, where the per-point feature changes equivariantly with the global transformation applied to the object. Part-level equivariant feature $F_i$ of each point $x_i$ changes equivariantly with the rigid transformation 
% $\in \text{SE(3)}$ 
applied to its parent part,
% itself, 
but remains invariant to transformations of other parts. 
We develop 
our network based on the Equivariant Point Network (EPN)~\cite{chen2021equivariant} with a novel pose-aware equivariant point convolution module to support part-level equivariance.
% to seek for such equivariant features. 
% The core of our 
% part-level equivariance 
% design
% design to achieve \textbf{part-level} equivariance 
% is a pose-aware equivariant point convolution module. 
In the following text, we would briefly review EPN, and then continue with our pose-aware equivariant point convolution. 

\vpara{Equivariant Point Network.}
EPN takes a point cloud $X$ containing $N$ points and a rotation group $G$ with $\vert G\vert$ elements as input and extracts $C$-dimensional per-point per-rotation features, forming a feature matrix $F\in \mathbb{R}^{N\times C\times \vert G\vert}$. $F$ is rotational and translational equivariant to a specific rigid transformation group $G_A$ induced by $G$.
% $G_A = \{ T_{A} = T_{\mathbf{t}} T_{g} \vert g\in G, \mathbf{t}\in \mathbb{R}^3\}$. 
The rotational equivariant transformation for each rotation element $g\in G$ in the feature domain is a corresponding matrix permutation of $F$ along the last dimension. 
The translational equivariance achieved by EPN is essentially translational invariance. Simply using relative point coordinates for convolution allows $F$ to remain the same while translating the input point cloud. %Please refer to EPN~\cite{chen2021equivariant} for more details.
\vpara{Pose-aware Equivariant Point Convolution.}
For part-level SE(3) equivariant features, we design a pose-aware point convolution strategy that operates on a point cloud with per-point poses.
% for feature communication. 
% includes per-point poses for feature  
% We design a pose-aware convolution strategy to arrive at this property. It operates on a point cloud and per-point poses for feature communication. 
% The key idea of the convolution is aggregating neighbours' features with their poses relative to the center point canceled. 
While conducting convolution within a group of points, our core idea is to align point poses to the pose of the group center. Since we use the pose of a point to refer to the pose of its parent part, such alignment could cancel out the influence of the varying articulation states on the geometric description of each point. Intuitively speaking, if a point comes from the same part as the group center, information will just get aggregated as a normal convolution. While a point comes from a different part from the group center, pose alignment will canonicalize the articulation state so that the convolution outcome remains the same regardless of the articulation state change. Our pose-aware convolution strategy allows aggregating context information from different parts but avoids feature changing as the articulation changes.
Equipping EPN with this strategy, we are able to achieve part-level equivariance since the feature of each point only changes as its parent part transforms but remains invariant to the transformation of other parts.
We then formally define our convolution operator. Taking a point cloud $X$ and the per-point pose 
% $P = \{  \mathcal{P}_i \vert 1\le i\le N \}$ 
$P = \{  {P}_i \vert 1\le i\le N \}$ 
% and the input point cloud $X$ 
as input, 
% designed 
% pose-aware point 
our convolution operator for the point $x_i$'s feature 
% vector 
at the rotation element $g$ is as follows: 
% the designed pose-aware point convolution strategy is as follows:
%%%%% some understandings %%%%%
% That could achieved by inversely transforming the $g$-rotated shape by the relative pose between each contextual part and the current part for each $g\in G_g$. 
% For an intuitive understanding, the estimated part pose could help us define a reference frame where the pose of each part is canonicalized. Then, the feature communication between parts invariant to part pose change could be enforced by communicating features in the reference frame. 
% Then, features equivariant to part $i$'s pose change can be learned based on the reference frame posed by the part pose equivariant to the part pose change. 
% Here, we just project the pose back to the input shape and achieve equivariance from a pose-induced manner to the geometry-induced way.
%%%%% some understandings %%%%%
% Formally, the designed pose-aware equivariant point convolution is as follows:
% feature communication (convolution) is as follows:
% \begin{align}
%   (\mathcal{F}* h_1)(x_i,g) = \sum_{\mathcal{P}_j^{-1}x_j\in \mathcal{N}_{x_i}^c} \mathcal{F}(x_j, g \mathcal{R}\mathcal{R}_j^{-1} ) h_1(g(x_i - \mathcal{P}\mathcal{P}_j^{-1}x_j)), \label{eq_part_level_convolution}
% \end{align}
\begin{align}
  (\mathcal{F}* h_1)(x_i,g) = \sum_{x_j\in \mathcal{N}_{x_i}} \mathcal{F}(x_j, g \mathbf{R}_i\mathbf{R}_j^{-1} ) h_1(g(x_i - {P}_{i}{P}_j^{-1}x_j)), \label{eq_part_level_convolution}
\end{align}
where $\mathcal{F}(x_i,g)$ is an input function, $h_1(x_i,g)$ is a kernel function, 
% where 
$\mathcal{N}_{x_i}$ is the set of points in the neighbourhood of $x_i$, $P_i$ and $P_j$
% $\mathcal{P}_i$ and $\mathcal{P}$ 
denote the input pose of point $x_i$ and point $x_j$ respectively, $\mathbf{R}_i$ and $\mathbf{R}_j$ is their rotation components.
% }
% respectively. 
% are the input rotations of point $x_i$ and point $x_j$ respectively.  
% \subsummary{Strategy}
% It takes point coordinates and estimated poses as input, further outputs per-point per-rotation point features. 
% Other notations are kept the same to those used in EPN~\cite{chen2021equivariant}.
% The part-level equivariance achieved by this strategy with ground-truth pose
% property 
% could be proved easily, which is deferred to the Appendix. 
% The part-level equivariant property of this convolution strategy with ground-truth pose could be proved easily 
% The part-level equivariant property can be achieved when the estimated per-point pose is accurate. 
% The property can be proved easily, which is deferred to the Appendix. 
% It can be proved easily, which is deferred to the Appendix. 
We prove that using the above convolution within EPN leads to part-level equivariance in the Appendix~\ref{sec_appen_part_level_equiv}.
% It is easy to prove the part-level equivariant property of this point convolution, which is deferred to the Appendix. 
% It is worth mentioning that 
% j
We highlight that 
% \textcolor{myblue}{
we adopt an iterative pose estimation strategy (see Appendix~\ref{sec_appen_additional_explanations} for details) for the per-point poses and rotations 
% Therefore, input per-point poses and rotations 
in Eq.~\ref{eq_part_level_convolution}, which are initialized to be identity in the first iteration. 
\subsection{Part Shape, Structure, and Pose Disentanglement}
% \subsection{Articulated Object Pose Estimation}
\label{sec_method_ssl_method}
To obtain a fine-grained understanding of an articulated object, we disentangle three types of information from the input: 
1) {Canonical shape}; 2) Object structure; 3) Articulated object pose.
% Articulated object pose.
% \textcolor{myblue}{
% Taking part-level SE(3) equivariant features as input, 
% % Based on part-level SE(3) equivariant features, 
% a self-supervised slot-attention module~\cite{locatello2020objectcentric} is first added to group points into parts. 
% Following the equivariant network, 
% a self-supervised slot-attention module~\cite{locatello2020objectcentric} first groups points into parts. 
To be more specific, 
% based on per-point features from extracted by our  part-level SE(3)-equivariant network from an input shape, we 
% from an input shape 
we first 
use the designed 
% design a 
part-level SE(3)-equivariant network to extract per-point features from an input shape. 
We then 
leverage a self-supervised slot-attention module to group the featured points, forming a set of featured parts for the disentanglement. 
% is first added to group points into parts. 
% is added for part segmentation. 
% We then use part-level features for the disentanglement. 
% by which part-level information 
% % we add a self-supervised slot-attention module~\cite{locatello2020objectcentric} for part segmentation, by which 
% % part-level information 
% could be leveraged for the following disentanglement. 
We predict a canonical shape for each part to induce the category-level canonical part spaces required by part pose definition.
% \textcolor{myblue}{
% Then we transform and combine them together based on predicted structure and poses gradually to the observed shapes. 
Then we disentangle structure and pose-related information that gradually transform canonical part shapes to the observed shape. 
First, we predict \emph{part-assembling parameters} to transform each canonical part shape to form the canonical object shape. 
% First, we transform each canonical part shape to form the canonical object shape via  \emph{part-assembling parameters}. 
After that, the \emph{kinematic chain}, \emph{joint parameters} and \emph{joint states} are predicted to articulate the canonical object shape into the observed articulation state. 
% After that, the canonical object shape is articulated to the observed articulation state via the \emph{kinematic chain}, \emph{joint parameters} and \emph{joint states}. 
Finally, a \emph{base part rigid transformation}  is predicted to further transform the  resulting articulated object to  the observed shape in the camera space. 
We will elaborate details of the above designs in the following text. 
\vpara{Part Proposal.}
% \todo{optionally...}
% \eric{With the part-level SE(3) equivariant backbone, we obtain per-point equivariant features.}
% \eric{To further extract part-level features, we introduce a part proposal module.}
% To better perform the part-by-part reconstruction task, we add a part proposal module for shapes with revolute parts for further part-level feature extraction. 
% \eric{This module leverages a grouping function to group input points into $K$ segments so that point features from the same segment can be pooled for part features.}
% The task of this module is to learn a point grouping function $\rho_g$ that maps per-point equivariant feature output by the feature backbone to $K$ clusters. 
% \eric{The grouping function should be SE(3) invariant. Specifically, we leverage an attention-pooling operation together with a slot attention module~\cite{locatello2020objectcentric} for the grouping purpose.}
% \eric{The attention pooling operation extracts per-point invariant features by pooling equivariant features corresponding to different rotation elements.}
% \eric{The slot attention module then consumes the per-point invariant features and learns a point-slot association matrix. Each slot corresponds to a part in our case.}
% With the part-level SE(3)-equivariant backbone, we obtain per-point equivariant features $F$. 
The part proposal module groups $N$ points in the input shape $X$ into $K$ parts for per-part equivariant features extraction. 
It learns an invariant grouping function that maps $X$ together with a point feature matrix $F$ to a point-part association matrix $\mathbf{W}\in \mathbb{R}^{N\times K}$. 
Specifically, we adopt an attention-pooling operation for the per-point invariant feature together with a slot attention module~\cite{locatello2020objectcentric} for the grouping purpose.
% The attention pooling operation extracts per-point invariant features $F_{inv} \in \mathbb{R}^{N\times C}$ by pooling equivariant features transformed by a per-point per-rotation MLP from $F\in \mathbb{R}^{N\times C\times \vert G_g\vert}$ corresponding to different rotation elements. 
% The slot attention module then consumes the per-point invariant features and learns a point-slot association matrix. Each slot corresponds to a part in our case.
Based on the proposed parts, we can group points in the input shape $X$ into $K$ point clouds $\{ X_i\vert 1\le i\le K \}$ and compute the per-part equivariant feature $\{ F_i\vert 1\le i\le K\}$.
\vpara{Shape: Canonical Part Shape Reconstruction.}
% For each part $i$, we learn to predict an invariant canonical shape $Z_i$ from its equivariant feature $F_i$.
% For canonical shape reconstruction, we want to learn an invariant module to reconstruct an invariant reference frame $Z_i$ for each part $i$ based on which the part pose could be defined. 
% Canonical shape reconstructions aims at reconstructing a category-level canonical part shape for each part. 
With per-part equivariant features, 
% and observed part shape, 
we aim to predict a canonical shape for each part which should be aligned within a certain category so that the category-level part pose can be defined.
% to help with the emergence of category-aligned canonical part spaces so that the category-level 
%pose-invariant  canonical part spaces of the shape collection.
% category-level part poses. 
% so that expectation for predicting category-level part poses. 
% \subsummary{Why}
% Canonical part shapes should be invariant to all parts’ rigid transformations. 
The canonical shape for each part should be invariant to every parts’ rigid transformations. 
% \subsummary{Overview}
% \subsummary{Property}
% The reconstructed part shape can be treated as the reference frame of the part, based on which the part pose could further be defined. 
% \subsummary{Purpose}
% Such reconstruction for each part requires invariance to all parts' rigid transoformations. 
% \subsummary{Property}
% Therefore, we choose to adopt the same SE(3)-invariant canonical shape reconstruction module as used in~\cite{li2021leveraging}.
Thus, we adopt an SE(3)-invariant canonical shape reconstruction module
% , which is 
constructed based on an SO(3)-PointNet module as utilized in~\cite{li2021leveraging}.
% To achieve this goal, we choose to adopt the same SE(3)-invariant canonical shape 
% Specifically, we adopt the same SE(3)-invariant canonical shape reconstruction module as used in~\cite{li2021leveraging}. 
The reconstruction module converts per-part equivariant features $F_i$ into per-part invariant features through attention pooling first and then predicts an SE(3)-invariant shape $Z_i$ for each part.

\vpara{Structure: Kinematic Chain Prediction.} 
In addition to the canonical shape of each part, 
we also need to understand the kinematic chain of a shape. 
The kinematic chain defines how different parts are connected and the order they get transformed when a cascaded transformation happens.
, \emph{i.e.} 
from chain leaves to the chain root. 
% chain root to chain leaves.
%We therefore use a part index sequence $(i_1,..., i_K)$ to encode the transformation order and use a pair set , where $1\le i_1,...,i_K \le K$. 
To estimate the kinematic chain for a given shape, we first construct an adjacency confidence graph from object parts and then extract its 
% a 
maximum spanning tree consisting of the set of confident adjacency edges. We set the part with the largest degree in the graph to be the root of the tree, which will also serve as the base part of the object.
% from it, 
% denoted as $\mathcal{T} = (\mathcal{V}, \mathcal{E}_{\mathcal{T}})$.
The transformation order is further predicted as the inverse DFS visiting order of the tree.
% The last part of the predicted kinematic chain is treated as the base part, \emph{i.e.} $i_K$.
% A kinematic chain defines the part adjacency relationship and part transformation order in the canonical object space, \emph{i.e.} $(i_1,..., i_K)$, where $1\le i_1,...,i_K \le K$ are the indices of parts. 
% \subsummary{Overview}
% It is important for joint parameters and joint states prediction. 
% the kinematic chain should be an invariant property, not affected by input objects’ articulation state variations. 
Notice the kinematic chain should not be affected by the articulated input pose, we therefore leverage per-part SE(3)-invariant features for estimation. 

\vpara{Structure: Joint Parameters Prediction.}
% \vpara{Invariant Kinematic Chain Prediction.}
% \todo{for category with more than two parts}
For each pair of adjacent parts
% $(i,j)\in \mathcal{E}_{\mathcal{T}}$
, we will then infer their joint parameters, 
% between them, 
including an invariant pivot point $\mathbf{p}_{i,j}^v$ and a joint axis orientation hypothesis $\mathbf{u}_i^g$ for each rotation element $g\in G_g$. 
% \subsummary{Overview and purpose}
% $\vert G_g\vert$ joint axis direction hypotheses  $\{\mathbf{u}_i^{g} \}$, where the upper footnote $g$ denotes the group element $g\in G_g$. 
% we predict an invariant pivot point $\mathbf{p}_{i,j}^v$ from their equivariant feature $\{ F_i, Fcor \}$ and coordinate matrices $\{ X_i, X_j \}$.
%%%%% %%%%%
% Firstly, we predict an invariant pivot point $\mathbf{p}_{i,j}^v$ from their equivariant feature $\{ F_i, F_j \}$ and coordinate matrices $\{ X_i, X_j \}$.
% Firstly, we still adopt an invariant shape reconstruction module to predict an invariant pivot point $\mathbf{p}_{i,j}^v$ for each adjacent parts $i,j$ from their equivariant feature $( F_i, F_j )$ and coordinate matrices $(X_i, X_j)$, considering the invariant property required for the pivot point prediction. 
% % using the same invariant function as that for canonical shape reconstruction, \todo{add notation}. 
% After that, we regress 
% % $\vert G_g\vert$ 
% joint axis directions $\{\mathbf{u}_i^{g} \vert g\in G_g \}$ 
% % for rotation matrix or translation scales and translation axis $\{ (\mathbf{u}_i^{g}, s_i^{g})\}$ 
% from each part's equivariant feature $F_i$.
% % for each $g\in G_g$.
% %  depending on the part type of $i$. 
%%%%% %%%%%
For pivot points, we treat them as invariant properties and still adopt an invariant shape reconstruction module for prediction. 
Specifically, we predict the pivot point $\mathbf{p}_{i,j}^v$ between every two adjacent parts $i,j$ from their equivariant feature $( F_i, F_j )$ 
% and the coordinate matrices $(X_i, X_j)$ 
using an invariant shape reconstruction module~\cite{li2021leveraging}. 
For joint axis orientations, we regress an axis orientation hypothesis $\mathbf{u}_i^g$ for part $i$ corresponding to each 
% for 
rotation group element $g\in G_g$ from its equivariant feature $F_i$. 

\vpara{Pose: Part-assembling Parameters Prediction.}
% Part-assembling parameters transform canonical part shapes $\{ Z_i \}$ from their canonical part spaces to the canonical object space. 
% Aiming for understanding part poses in the object-level with part associations and part kinematics well considered and modeled, we propose to construct a canonical object shape by assembling parts together. 
% % each part canonical reconstruction $\{ Z_i \}$ from their canonical part spaces to the canonical object space.
% \subsummary{Purpose}
% We choose to assemble parts together via rigid transformations. The predicted transformation for each part should be invariant to all parts’ rigid transformations. 
% \subsummary{Property}
Part-assembling parameters transform the predicted canonical part shapes to assemble a canonical object shape.
% such that we can get the canonical object shape
% canonical object space 
% by assembling the transformed parts together. 
% The frame of the assembled canonical object shape is then treated as the canonical object space. 
% from canonical part spaces to the canonical object space. 
As parameters connecting invariant canonical shapes, they should be 
% invariant to object articulation variations as well. 
invariant to every parts’ rigid transformations as well. 
% Here, we simply predict an offset term $\mathbf{p}_i^c \in \mathbb{R}^{3}$ for each part $i$ that acts as a translation vector between two shapes.
Here, we simply predict a translation vector $\mathbf{p}_i^c \in \mathbb{R}^{3}$ for each part $i$. 
We predict them through invariant shape reconstruction modules from per-part equivariant feature $\{F_i\vert 1\le i\le K\}$.
% We treat it as an invariant shape reconstruction problem and predict it from per-part equivariant feature $\{F_i, 1\le i\le K\}$. 
% connecting two spaces.
% that transforms the part from the canonical part space to the canonical object space. 
% The canonical object shape is then formed by assembling those transformed part shapes together.
% \subsummary{Purpose}
% to assemble together for the canonical object shape 
% Based on part canonical reconstructions $\{ Z_i \}$, we assemble those parts together by predicting per-part transformation for part assembling. 
% We bridge the predicted the canonical part space with canonical object space by predicted per-part transformations. 
% Therefore, we simply predict an invariant central point $\mathbf{p}_i^c$ performed as a translation vector for each $i$ depicting its location in the canonical object space. 
% Specifically, we view it as an invariant shape reconstruction problem and solve it via an SE(3)-invariant shape reconstruction module~\cite{li2021leveraging} that operates on per-part equivariant feature $\{F_i, 1\le i\le K\}$.
% The canonical object shape is formed by assembling parts together $Z = \{ Z_i + \mathbf{p}_i^c \vert 1 \le i\le K \}$.
% \textcolor{myblue}{
We can then assemble predicted canonical part shapes together to form the canonical object shape: $Z = \{ Z_i + \mathbf{p}_i^c \vert 1 \le i\le K \}$.
% }
% We can then assemble predicted canonical part shapes together to the canonical object shape: $Z = \{ Z_i + \mathbf{p}_i^c \vert 1 \le i\le K \}$.

% \vpara{Base Part 6D Pose Prediction.}
% \vpara{Articulated Object Pose Prediction.}
\vpara{Pose: Joint States Prediction.}
% Based on the predicted kinematic chain, 
% With the predicted part structure
%%%%% %%%%%
% Joint state of a part describes the rotation angle or the translation scalar of each part's transformation along the joint. 
% \subsummary{Overview}
% We choose to predict joint states from each part's equivariant feature to reduce the optimization difficulty. 
% Specifically, for each part $i$, we predict a joint state (a rotation angle $\theta_i^g$ for revolute parts or a translation scalar $s_i^g$ for prismatic parts) for each rotation element $g\in G_g$ from its equivariant feature $F_i$. 
% \subsummary{Approach}
%%%%% %%%%%
Joint states describe the articulation state of an object. 
% equivariant feature %
For each part $i$, we predict a joint state hypothesis for each rotation element $g\in G$ from its equivariant feature $F_i$, \emph{i.e.} a rotation angle $\theta_i^g$ for a revolute part or a translation scalar $s_i^g$ for a prismatic part. 
% \textcolor{myblue}{
We can therefore articulate the canonical object shape based on the predicted kinematic chain and joint states with the base part fixed, 
so as to match the object articulation from the input observation.

\vpara{Pose: Base Part Rigid Transformation.}
The base part rigid transformation needs to transform the articulated canonical object shape to the camera space. Since we have previously predicted joint states hypotheses for all rotation element $g$, we will also need multiple base transformation hypotheses correspondingly. We simplify the base part transformation to be a rotation, which proves to be effective in practice. A straightforward way is to use the rotation matrix corresponding to each rotation element $g$ as the base transformation hypothesis. We follow this idea but also predict an additional residual rotation as a refinement.
% \textcolor{myblue}{
By transforming the articulated shape of the canonical object shape via the predicted base part rigid transformation, we can align the resulting shape with the observed input object.
% } 
% We postpone the determination of a unique transformation but instead compute multiple transformation hypotheses. Specifically, we consider the rotation matrix corresponding to each rotation element $g$ as one rotation hypothesis and predict an additional residual rotation and translation term to further refine the hypothesis.\eric{what about translation?} 
% \subsummary{Overview}
% The base part is defined as the last part $i_K$ of the predicted kinematic chain. 
% We predict a residual rotation for the discrete major rotation matrix corresponding to each rotation element $g$. 
% corresponding to the rotation element $g$. 
% Each residual translation vector is set to zero. 
%%%%%% %%%%%%
% Then, we predict a rotation  matrix $\Delta \mathbf{R}_{i_K}^g$ for each group element $g\in G_g$ from its equivariant feature $F_{i_K}$. 
% % For base part rigid transformation, we first define a base part  as the last part $i_K$ of kinematic chain. 
% % Then, we predict $\vert G_g\vert$ rotation matrices $\{ \Delta \mathbf{R}_{i_K}^{g} \}$ from its equivariant feature $F_{i_K}$. 
% After that, the hypotheses for the base part's rigid transformation are like $\{ (\mathbf{R}^{g} \Delta\mathbf{R}_{i_K}^{g}) , \mathbf{1} \vert g\in G_g \}$, where $\mathbf{R}^g$ is the discrete major rotation matrix of the rotation element $g$.  
%%%%%% %%%%%%
% \subsummary{Approach}

\vpara{Articulated Object Pose.} 
% The articulated object pose can further be calculated based on the above predicted information. 
% \subsummary{Overview}
% Firstly, the rigid transformation of each part $i$ from its canonical part space to the canonical object space is set to its predicted central point $\mathbf{p}_i^c$. 
% Then, the articulated rigid transformation hypotheses of each part $i$ in the canonical object space is calculated 
% % by 
% based on 
% the predicted kinematic chain, joint parameters and joint states. 
% % We denote the predicted hypotheses for each part $i$ as $\{ (\mathbf{R}_i^{g,o}, \mathbf{t}_i^{g,o}) \vert g\in G_g \}$, where $o$ represents the canonical object space. 
% % $o$ denotes the canonical object space. 
% At last, we take the predicted hypotheses for the base part's rigid transformation as the part-common rigid transformation hypotheses.
% % part-common transformation hypotheses. 
% Part pose hypotheses for each part $i$ from its canonical part space to the camera space can be calculated as are denoted as $\{ P_i^g = (\mathbf{R}_i^g, \mathbf{t}_i^g) \}$. 
% % $\{ (\mathbf{R}_i^g = \Delta \mathbf{R}_{i_K}^{g} \mathbf{R}_i^{g,o}, \mathbf{t}_{i}^g = \Delta \mathbf{R}_{i_K}^{g} \mathbf{t}_i^{g,o}) \}$. 
% \subsummary{Approach or components}
With the above predicted quantities, we can calculate per-rotation
articulated object pose hypotheses for an input articulated object $X$, including three parts: 
% % to each rotation group element $g$.
% % for an input articulated object. 
% Specifically, the per-rotation articulated pose 
% % corresponding to each rotation element $g\in G$ 
% is composed of three parts: 
% % The articulated pose of an articulated object is composed of three parts: 
1) translation $\mathbf{p}_i^c$ of each part $i$ which assembles category-aligned canonical parts into a canonical object; 2) per-rotation articulated transformation of the canonical object based upon the predicted kinematic chain, joint parameters and per-rotation joint states; 3) per-rotation base part rigid transformation which transforms the articulated canonical object into the camera space. 
% The rigid transformation hypotheses corresponding to each rotation element in the group $G_g$ for each part $i$ from the canonical part space to the camera space are denoted as  $\{  P_i^g = (\mathbf{R}_i^g, \mathbf{t}_i^g) \vert g\in G_g \}$. 
The rigid transformation hypothesis for each part $i$ corresponding to each rotation element $g\in G$
% in the rotation group $G$ 
is denoted as $P_i^g = (\mathbf{R}_i^g, \mathbf{t}_i^g)$. 
We treat them as part pose hypotheses. 

\subsection{Shape Reconstruction-based Self-supervised Task}
% \subsection{The Self-supervised Task}
\label{sec_method_self_supervised_task}
% \todo{something for the task...}
Based on the reconstructed canonical part shapes and predicted per-rotation part pose hypotheses, we can get per-rotation shape reconstruction for each part $i$: 
% the per-rotation reconstructed shape for each part $i$: 
$\{ Y_i^{g} = \mathbf{R}_i^g Z_i + \mathbf{t}_i^g \vert g\in G\}$. 
% \subsummary{Part reconstruction}
% Based on the predicted per-part per-rotation part pose, we can get per-rotation reconstructed shape for each part $i$: $\{ Y_i^{g} = \mathbf{R}_i^g Z_i + \mathbf{t}_i^g \}$. 
% Based on the per-rotation reconstructed shape, 
% Then based on those reconstructed part shapes, we use a shape reconstruction task to self-supervise the network's learning process. 
% Besides, some regularizations are added for predicted joints. 
% joint prediction. 
% Further, we adopt a part-by-part reconstruction task to self-supervise the network.
A part-by-part reconstruction task is adopted to self-supervise the network. 
% 's learning. 
Besides, 
we add a regularization term for each predicted joint so that the joint indeed connects two parts. 
% some regularization terms are added for predicted joints. 

% \zz{per-rotation?}

\vpara{Shape Reconstruction-based Self-supervised Loss.}
% Using the method proposed in~\cite{li2021leveraging}, we first estimate the pose $(\mathbf{R}_{glb}, \mathbf{t}_{glb})$ of the input shape $X$ by  treating it as a rigid object for better performance.
% Together with $(\mathbf{R}_{glb}, \mathbf{t}_{glb})$, we can get a 6D pose hypothesis for each part $i$ for each 
% $g_j \in G_g$ from the proposed invariant canonical space to the observed camera space: 
% $ \{ P_i^j  = (\mathbf{R}_{glb} \mathbf{R}_g^j \mathbf{R}_i^j, \mathbf{R}_{glb}\mathbf{R}_g^j \mathbf{t}_i^j)\} $.
% $ \{ P_i^j \} \in \mathbb{R}^{(9 + 3) \times \vert G_g\vert}$. 
% $g_j\in G_g$: $\{ P_i^j = (\mathbf{R}_{glb} \mathbf{R}_g^j \mathbf{R}_i^j, ) \}$
% todo: xxx
% After we have got $G_g$ part pose hypotheses for each part, 
% Then 
The per-rotation shape reconstruction for the whole object can be calculated by concatenating all part reconstructions: 
$Y^{g} = \{ Y_i^{g} \vert 1\le i\le K \}$.
% $Y^{g} = \{ Y_i^{g} \}_{1\le i\le K}$. 
% \subsummary{Per-rot reconstruction}
% For each $g \in G_j$, the whole shape can be reconstructed by concatenating all part reconstructions: 
% we can get the reconstructed whole shape by concatenating all part reconstructions: 
% $ Y^{g} = \{ Y_i^{g} \}_{1\le i\le K}$.
% $Y^j = \bigcup_{1\le i \le K}\{ Y_i^j = \mathbf{R}_i^j Z_i + \mathbf{t}_i^j \}$. 
We then adopt a min-of-N loss between the input observation $X$ and the reconstructed posed point clouds: 
\begin{align}
    \mathcal{L}_{rec} = \min_{ g \in G } d(X, Y^{g}),  \label{eq_recon_loss}
\end{align}
where $d: \mathbb{R}^{N_X\times 3} \times \mathbb{R}^{N_Y\times 3}\rightarrow \mathbb{R}$ denotes the distance function between two point clouds and could be unidirectional or bidirectional Chamfer Distance as an example. 

\vpara{Regularization for Joint Prediction.}
% \subsection{Constraints for Joint Prediction} \label{sec_method_joint_prediction}
% One challenging thing of part motion modeling in the canonical space is how to predict the joint $\mathbf{u}_{i,j}$ between two parts $\mathbb{S}_i, \mathbb{S}_j$. 
% In this method, we predict the joint between two adjacent parts $i,j$ by predicting the joint direction $\mathbf{u}_{i}$ and a pivot point $\mathbf{p}_{i,j}^v$. 
% Predicted joints should behave like real joints for 1) predicting ``real’’ part articulated transformations in the canonical object space, and 2) part-level equivariant features that are more likely output by the pose-aware point convolution module. 
Predicted joints should connect adjacent parts and support natural articulations.
However, just supervising joint parameters from the reconstruction loss is not sufficient for the needs above. 
Therefore, we devise a point-based joint constraint term
% for predicted joints 
% add constraints on the predicted joint 
for each predicted joint $(\mathbf{u}_i^{g_0}, \mathbf{p}_{i,j}^v)$, 
% $\{(\mathbf{u}_i^{g_0}, \mathbf{p}_{i,j}^v)\}$, 
where $g_0 = \text{argmin}_{g\in G}d(X, Y^g)$ (Eq.~\ref{eq_recon_loss}).
% is the index of the selected reconstruction hypothesis that achieves the minimum distance value $d(X, Y^{g_0})$. 
% \subsummary{Method overview}
%%%% move to supp %%%%
% It is based on the assumption that the joint between two parts $i,j$ should go through the two adjacent part shapes connected by it both before and after the articulation transformation in the canonical space. 
%%%% move to supp %%%%
% \subsummary{Assumption}
Specifically, given the predicted pivot point $\mathbf{p}_{i,j}^v$ and joint orientation $\mathbf{u}_i^{g_0}$, we independently randomly sample a set of points from the joint by shifting the pivot point $\mathbf{p}_{i,j}^v$: $P_{i,j}^v = \{ \mathbf{p}_{i,j}^{v,k} \vert 0\le k\le K^v\}$. 
%%%%%%%%%%%%%%%
% offset scales from a uniform distribution: $\{ s_k \vert s_k \sim \mathcal{U}(a,b), 1\le k\le K^v \}$, where $ a\in \mathbb{R} < b \in \mathbb{R}$ are hyper-parameters that can be tuned for each category. 
% After that, a set of shifted pivot points are generated via $\{ \mathbf{p}_{i,j}^{v,k} = \mathbf{p}_{i,j}^v + s_k \cdot \mathbf{u}_i^{g_0} \vert 1\le k\le K^v \}$, where $K^v$ is the number of sampled shifted pivot points. 
% Together with the pivot point, also denoted as $\mathbf{p}_{i,j}^{v,0}$, we get the (shifted) pivot point set $P_{i,j}^v = \{ \mathbf{p}_{i,j}^{v,k} \vert 0\le k\le K^v\}$.
%%%%%%%%%%%%%%%
% where $K^v$ is the number of sampled shifted pivot points. 
% Loss for joint constraints are added as follows:
The joint regularization loss term is as follows: 
% Denote shapes of part $i$ in the canonical object space before and after the articulated transformation as $Z_i^1$ and $Z_i^2$ respectively,
% % Assume shapes of part $i$ in the canonical object space before and after the articulated transformation are denoted as $Z_i^1$ and $Z_i^2$ respectively, 
% the loss for joint regularization is designed as follows: 
\begin{align*}
    \mathcal{L}_{reg} = \sum_{(i,j)\in \mathcal{E}_{\mathcal{T}}} d(P_{i,j}^v, Z_i^2) + d(P_{i,j}^v, Z_j^2) + d(P_{i,j}^v, Z_i^1) + d(P_{i,j}^v, Z_j^1),
\end{align*}
% \begin{align*}
%     \mathcal{L}_{reg} = \sum_{(i,j)\in \mathcal{E}_{\mathcal{T}}} d(P_{i,j}^v, \mathbf{R}_i^{g_0,o} Z_i' + \mathbf{t}_i^{g_0,o}) + d(P_{i,j}^v, \mathbf{R}_j^{g_0,o} Z_j' + \mathbf{t}_j^{g_0,o}) + d(P_{i,j}^v, Z_i') + d(P_{i,j}^v, Z_j'),
% \end{align*}
where $Z_i^1$ and $Z_i^2$ are shapes of the part $i$ in the canonical object space before and after its articulated transformation, 
$\mathcal{E}_{\mathcal{T}}$ is the set of adjacent parts,  $d(X_1, X_2)$ is the unidirectional Chamfer Distance function from point cloud $X_1$ to $X_2$.
% \subsummary{Approach}
% $j_0$ is the index of the selected pose hypothesis that achieves the minimum distance value $d(X, Y^{j_0})$. 

Our final self-supervised shape reconstruction loss is a linear combination of the above two loss terms: 
% Thus, the total loss for the self-supervised shape reconstruction task is 
$\mathcal{L} = \mathcal{L}_{rec} + \lambda \mathcal{L}_{reg}$, where $\lambda$ is a hyper-parameter. 
% \subsummary{Summary for loss}

% \zz{??}

%%% move to supp? %%%%
% \vpara{Iterative Pose Estimation.}
% It can be noticed that the designed part-level equivariant network needs estimated per-point pose as input. 
% The ideal part-level equivariance can only be achieved when the input per-point pose is accurate. 
% Therefore, we design an iterative pose estimation approach where the estimated pose from the previous iteration is taken as the input per-point pose in the current iteration. 
% We wish that the pose estimation quality can be improved across such iterations. 

\begin{figure*}[ht]
    \centering
      \includegraphics[width=0.80\textwidth]{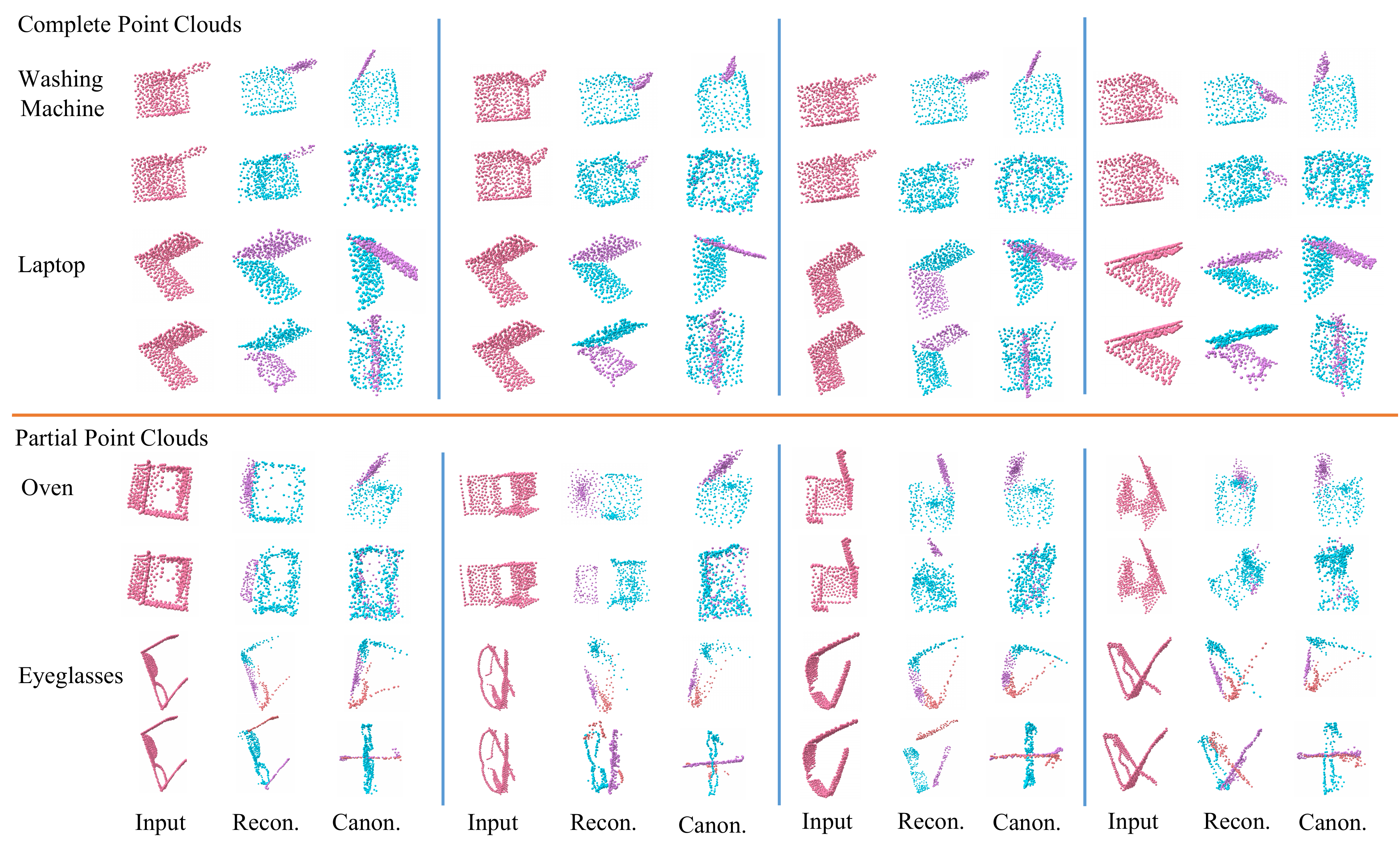}
      \vspace{-5pt}
    \caption{ \footnotesize
        Visualization for qualitative evaluation. 
        % \textcolor{myblue}{
        For every two lines, the first line draws the results of our method, and the second line draws those of NPCS. 
        % For ours, e
        Every three shapes from the left side to the right side are the input point cloud (\textbf{Input}), reconstruction (\textbf{Recon.}), and the reconstructed canonical object shape (\textbf{Canon.}).
        % For NPCS, every three shapes are the input point cloud (\textbf{Input}), reconstruction (\textbf{Recon.}), and the reconstructed canonical part shapes (\textbf{Canon.}) from left to right. 
        % Since NPCS does not predict shape reconstructions explicitly, we take parts transformed from their canonical part spaces to the camera space as their reconstructions. 
        % Not that NPCS does not output shape reconstructions. Therefore we take parts transformed from the canonical part spaces to the camera space as their reconstructions. 
        % \textbf{We align input shapes and reconstructions across different instances just for a better view.} 
        \textbf{We do not assume input shape alignment but align them here when drawing just for a better view.} Please zoom in for details. 
        % For partial Oven objects, input shapes and reconstructions from those two methods are rotated to display their partial patterns. In practice, their global poses may vary from each other. 
        }
    % }
    \label{fig_canon_vis}
    % \vspace{-5pt}
    \vspace{-20pt}
\end{figure*}

% !TEX root = ../main.tex

\section{Experiments}

% \begin{itemize}
%     \item Complete/Partial Point Cloud.
%     \item With/Without Joint Modeling.
%     \item Single/Double Iteration(s).
%     % \item 
% \end{itemize}

% \textcolor{orange}{Complete pcs v.s. Depth pcs (global pose variation).}

% data preparation ---> add random rotations to the generated shape for data generation...
% 10 random rotations is added; 100 articulation samples; split the shape into 9:1 by shape indexes.
% datasets: drawer, 

% \todo{overview}

We evaluate our method on the category-level articulated object pose estimation task (sec.~\ref{sec_exp_pose_est}) to demonstrate its effectiveness. 
Besides, we also test its performance on two side tasks that can be completed by our network at the same time, namely part segmentation (sec.~\ref{sec_exp_seg}), and shape reconstruction (sec.~\ref{sec_exp_recon}). 
% \textbf{Quantitative evaluations on partial point clouds are deferred to the Appendix~\ref{sec_appen_exp_partial} to spare space.}
% and the priority of some designes. 
% to perform category-level part pose estimation without annotations, the priority of pose-aware part co-segmentation for articulated objects, especially when input shapes are not well alinged, as well as the priority of part-by-part shape reconstruction, compared with reconstructing shapes as single rigid objects. 

% In the following sections, we first introduce the data perparation process, and then results together with further analysis on the above three tasks. \textbf{To spare space, apart from some visualizations drawn in Figure~\ref{fig_canon_vis}, we present the experimental results on rendered depth point clouds in the Appendix}.
% Note that 

\subsection{Datasets}
Following previous literature~\cite{li2020category}, we choose seven categories from three datasets for evaluation on both complete shapes and rendered partial point clouds: 
% We evaluate the proposed method on shapes from three datasets for both complete shapes and rendered depth point clouds: 
1) Four categories from the Part-Mobility~\cite{wang2019shape2motion} dataset,  
% Shape2Motion~\cite{wang2019shape2motion} dataset,  
namely Oven, Washing Machine, Laptop (denoted as Laptop (S)), and Eyeglasses with  revolute parts.
% . Shapes in those categories only contain revolute parts. 
2) 
One category, Drawer with prismatic parts, from SAPIEN dataset~\cite{thomas2011one}.
% One category from SAPIEN~\cite{thomas2011one} dataset, namely Drawer with prismatic  parts.
% Shapes in this dataset only contain prismatic parts. 
3) Two categories from a real dataset  HOI4D~\cite{liu2022hoi4d}, namely Safe and Laptop (denoted as Laptop (R)) with revolute parts.
% Shapes in those categories only contain revolute parts. 
%%%% Appendix %%%%
% For all of those categories, we split shapes into train/test sets by a 9:1 ratio. 
% For each shape, we generate 100 posed shapes in different articulation states. 
% For complete point clouds, we randomly sample 10 global rotations from SO(3) space to generate globally posed shape. 
% For depth point clouds, we render depth images of complete object instances using the same rendering method described in~\cite{li2021leveraging}. 
% The difference is that we manually set a view point range for each category to ensure that all parts are visible in the rendered depth images. 
%%%% Appendix %%%%
Please refer to the Appendix~\ref{sec_appen_data_prepare} for data preparation details. 
% more details w.r.t. data praparation to the Appendix.
% The difference is that we manually set a range for valid view points to ensure that all parts are visible in the rendered depth images. 
% view point range for each category to ensure that all parts are visible in the rendered depth images. 
% The difference is that we manually set a view range for each category to ensure that all parts are visible in the rendered depth images. 
% Details for shape generation are deferred to the 
% \todo{Details?}

\subsection{Category-level Articulated Object Pose Estimation} \label{sec_exp_pose_est}
\vpara{Metrics.}
% We test the performance of the method on 1) part pose estimation; 2) joint parameters estimation.
Following ~\cite{li2020category}, we use the following metrics to evaluate our method: 1) Part-based pose-related metrics, namely per-part rotation error $R_{err}(^{^\circ})$ and per-part translation error $T_{err}$, both in the form of mean and median values; 2) Joint parameters, namely joint axis orientation errors $\theta_{err}(^\circ)$ in degrees and joint position error $d_{err}$, both in the form of mean values. 
% we test on two sub-tasks: 1) part pose estimation; 2) joint parameters estimation.
% Part pose estimation metrics are the same as those used in~\cite{li2020category}, namely rotation error $R_{err}(^{^\circ})$, translation error $T_{err}$ in the form of mean and median values, angle error $\theta_{err}(^\circ)$ and distance error $d_{err}$ for predicted joints in the form of mean values. 
% We adopt the same method as that used in~\cite{li2021leveraging} to evaluate the performance of our method on category-level part pose estimation. 
Please refer to the Appendix~\ref{sec_appen_eval} for details of our evaluation strategy. 
% of 
% category-level part pose 
% estimation 
% evaluation strategy. 

\vpara{Baselines.}
Since there is no previous works that have exactly the same setting with ours, we choose NPCS~\cite{li2020category}, a \textbf{supervised} pose estimation method for articulated objects, and ICP, a traditional pose estimation approach, 
% for rigid objects, 
% with ground-truth part segmentation labels
% but with some modifications to estimate part pose, 
as our baseline methods.
To apply them on our articulated objects with arbitrary global poses, we make the following modifications: 
1) We change the backbone of NPCS to EPN~\cite{chen2021equivariant} (denoted as ``NPCS-EPN'') and add supervision on its discrete rotation mode selection process to make it work on our shapes with arbitrary global pose variations. 
We do observe that the NPCS without EPN will fail to get reasonable results on our data (see Appendix~\ref{sec_appen_more_exp_additional} for details). 
Beyond part poses, we also add a joint prediction branch for joint parameters estimation. 
% joint parameter prediction branch for joint axis orientation and position estimation. 
2) We equip ICP with ground-truth segmentation labels (denoted as ``Oracle ICP'') and register each part individually for part pose estimation. 
Notice that Oracle ICP cannot estimate joint parameters. 

\vpara{Experimental Results.} 
Table~\ref{tb_exp_pose_cmp} presents the experimental results of our method and baseline methods on complete point clouds. 
We defer the results on partial point clouds to Table~\ref{tb_exp_pose_cmp_partial} in the Appendix~\ref{sec_appen_exp_partial}. 
% Results on partial point clouds are deferred to Table~\ref{tb_exp_pose_cmp_partial} in the Appendix~\ref{sec_appen_exp_partial}. 
% It can be seen that, 
We can make the following observations:
% 1) 
% % Our method 
% Without any human label available during training, our method can outperform the Oracle ICP  with ground-truth segmentation labels by a large margin on all categories. 
% We can achieve the average rotation/translation estimation error 7.90$^{\circ}$/0.065, much better than those of Oracle ICP 
% As an unsupervised method, our strategy can achieve reasonable part pose estimation results on all tested categories, and even sometimes outperform the supervised NPCS, such as 
% Though our method could not outperform the supervised method on most metrics in those tested categories,
% % most of those categories, 
% it can still achieve relatively good and reasonable results,
% % on all of those categories 
% and even outperforms supervised NPCS on some metrics, like 
% rotation errors on Laptop (R).
% and joint parameter estimation errors on Safe. 
% Moreover, the method can even outperform the supervised method NPCS on some metrics in several categoreis, such as rotation errors on Laptop$^2$. 
1) 
As a self-supervised strategy, our average and per-category performance are comparable to that of the supervised baseline NPCS-EPN. 
We can even sometimes outperform NPCS-EPN 
% on some specific categories 
such as the joint axis orientation estimation on Safe. 
2) 
% Our method 
Without any human label available during training, our method can outperform the Oracle ICP  with ground-truth segmentation labels by a large margin in all categories. 
% 2) 
% % The performance of ICP is terrible, though with ground-truth segmentation labels when doing registerations. 
% It may be surprising that ICP performs terribly on those tested categories, even if it directly registers ground-truth parts. 
% % we directly register ground-truth parts. 
% % ICP, though with ground-truth segmentation labels when doing registerations, performs terrible on tested categories. 
% It may be caused by two reasons: local optimum encountered during ICP's optimization and the pose ambiguity resulted from part-symmetry. 
% % that ICP tends to get stuck in, and the pose ambiguity caused by part-symmetry. 
% The part-symmetry problem may make the ICP's optimization even harder. 
% Apart from the local optimum that ICP tends to get stuck in,  
% the pose ambiguity caused by part symmetry may be another reason for its 
% part symmetry-related ambiguity is another reason for its 
% unsatisfactory results.  
% pose ambiguity caused by registering symmetric parts independently is another reason for the poor performance.
% part symmetry is another reason for its poor performance since it will cuase some ambiguity in the pose estimation result. 
% We can see... \todo{xxx}
% \todo{about ICP?}
As a further discussion, the poor performance of Oracle ICP may be caused by the part-symmetry related problem. 
It would add ambiguity on part poses especially when we treat each part individually for estimation. 
Please refer to Appendix~\ref{sec_appen_symmetric_parts} for more discussions.
% on symmetric parts. 
For a qualitative evaluation and comparison, 
% In Figure~\ref{fig_canon_vis}, 
we visualize the input objects, reconstructions, and the predicted canonical object shapes
% predicted canonical reconstructions, and transformed reconstructions 
by our method and NPCS 
% on four categories 
in Figure~\ref{fig_canon_vis}. 
% Shapes drawn above the orange line are results for complete point clouds, while those drawn below are partial point clouds. 
% where shapes drawn above the orange line are with complete point clouds as input and those below the line with partial input point clouds. 
% From the figure, we can get a glimpse into the ability of the designed method to reconstruct the invariant canonical reference frame. 
Our method is able to reconstruct category-level aligned
% cross-instance aligned 
% invariant 
canonical shapes, which serve as good support for estimating category-level articulated object poses. 
% reference frames, which are crucial to predict category-level articulated object poses. 
% for predicting category-level articulated object poses. 
% The reconstructed canonical reference frame for objects in the Eyeglasses category are not aligned very 
% \todo{about eyeglasses}
% \todo{ICP}

\begin{table}[t]
    \centering
    \caption{\footnotesize 
    Comparison between the part pose estimation performance of different methods on all categories. 
    ``R'' denotes rotation errors with the value format  
    % , whose values are 
    % % presented for rotation errors 
    % in the format of 
    ``Mean $R_{err}$/Median $R_{err}$''. ``T'' denotes translation errors with the value format 
    % , whose values 
    % % presented 
    % are in the format of 
    ``Mean $T_{err}$/Median $T_{err}$''. ``J'' denotes joint parameters estimation results with the value format ``Mean $\theta_{err}$/Mean $d_{err}$''.
    ``Avg.'' refers to ``Average Value''.
    % ,whose values 
    % % Values presented 
    % are in the format of ``Mean $\theta_{err}$/Mean $d_{err}$''. 
    \textbf{Since Oracle ICP could not predict joint parameters}, we only present joint parameter prediction results of our method and supervised NPCS-EPN. 
    % only the results of supervised NPCS and our method on joint prediction are presented.
    For all metrics, the smaller, the better. 
    Best values and \textbf{Bold}, while second best ones are shown in \emph{\textcolor{blue}{blue}}. 
    } 
%   \vspace{-8pt}
	\resizebox{0.7\linewidth}{!}{%
    \begin{tabular}{@{\;}c@{\;}|c|c|c|c|c|c|c|c|c@{\;}}
    \midrule
        \hline
        \specialrule{0em}{1pt}{0pt} 
        ~ & Method & Oven & \makecell[c]{Washing \\ Machine} & Eyeglasses & Laptop (S) & Safe & Laptop (R) & Drawer & Avg. \\ 
        \cline{1-10} 
        \specialrule{0em}{1pt}{0pt}
        
        % mean error / Median error
        \multirow{6}{*}{R} & \makecell[c]{NPCS-EPN \\ (supervised)} & \makecell[c]{\textbf{5.47}/\textbf{4.45},  \\ \emph{\textcolor{blue}{7.35}}/\emph{\textcolor{blue}{7.30}}} & \makecell[c]{\textbf{4.76}/\textbf{4.07}, \\ \textbf{6.66}/\textbf{5.41}} & \makecell[c]{\textbf{2.75}/\textbf{2.44}, \\ \textbf{9.34}/\textbf{7.64}, \\ \textbf{7.93}/\textbf{6.74}} & \makecell[c]{\textbf{6.72}/\emph{\textcolor{blue}{6.08}}, \\ \emph{\textcolor{blue}{15.96}}/\emph{\textcolor{blue}{13.91}}} & \makecell[c]{\textbf{1.75}/\textbf{1.59}, \\ \textbf{2.67}/\textbf{2.50}} & \makecell[c]{\emph{\textcolor{blue}{8.20}}/\emph{\textcolor{blue}{7.12}}, \\ \emph{\textcolor{blue}{5.13}}/\emph{\textcolor{blue}{4.72}}} & \makecell[c]{\textbf{1.52}/\textbf{1.31}, \\ \textbf{2.01}/\textbf{1.81}, \\ \textbf{2.15}/\textbf{1.81}, \\ \textbf{1.14}/\textbf{0.94}} & \textbf{5.38}/\textbf{4.70}
        \\ \cline{2-10} 
        \specialrule{0em}{1pt}{0pt}

        ~ & Oracle ICP & \makecell[c]{{46.46}/38.56,  \\ {47.11}/43.41} & \makecell[c]{55.12/50.42, \\ 52.38/51.57} & \makecell[c]{34.41/23.25, \\ 34.58/25.82, \\ 35.71/25.12} & \makecell[c]{43.26/42.02, \\ 44.04/43.64} & \makecell[c]{52.80/56.02,\\ 53.04/52.13} & \makecell[c]{42.50/43.06, \\ 42.06/39.25} & \makecell[c]{50.15/47.14,\\50.12/47.15,\\50.11/47.15,\\50.07/46.41} & 46.11/42.48
        \\ 
        \cline{2-10} 
        \specialrule{0em}{1pt}{0pt}

        % mean error / Median error
        ~ & Ours & \makecell[c]{\emph{\textcolor{blue}{7.74}}/\emph{\textcolor{blue}{7.35}},  \\ \textbf{4.07}/\textbf{3.97}} & \makecell[c]{\emph{\textcolor{blue}{7.49}}/\emph{\textcolor{blue}{7.37}}, \\ \emph{\textcolor{blue}{19.27}}/\emph{\textcolor{blue}{19.19}}} & \makecell[c]{\emph{\textcolor{blue}{8.16}}/\emph{\textcolor{blue}{8.21}}, \\ \emph{\textcolor{blue}{12.29}}/\emph{\textcolor{blue}{10.89}}, \\ \emph{\textcolor{blue}{12.53}}/\emph{\textcolor{blue}{9.88}}} & \makecell[c]{\emph{\textcolor{blue}{7.34}}/\textbf{5.16}, \\ \textbf{10.41}/\textbf{9.34}} & \makecell[c]{\emph{\textcolor{blue}{9.03}}/\emph{\textcolor{blue}{9.09}}, \\ \emph{\textcolor{blue}{13.83}}/\emph{\textcolor{blue}{13.59}}} & \makecell[c]{\textbf{5.71}/\textbf{3.61}, \\ \textbf{3.64}/\textbf{2.84}} & \makecell[c]{\emph{\textcolor{blue}{3.18}}/\emph{\textcolor{blue}{2.73}}, \\ \emph{\textcolor{blue}{3.18}}/\emph{\textcolor{blue}{2.73}}, \\ \emph{\textcolor{blue}{3.18}}/\emph{\textcolor{blue}{2.71}}, \\ \emph{\textcolor{blue}{3.18}}/\emph{\textcolor{blue}{2.71}}}  & \emph{\textcolor{blue}{7.90}}/\emph{\textcolor{blue}{7.14}}
        \\ \cline{1-10} 
        \specialrule{0em}{1pt}{0pt}
        
        \multirow{6}{*}{T} & \makecell[c]{NPCS-EPN \\ (supervised)} & \makecell[c]{\textbf{0.029}/\textbf{0.029}, \\ {0.020}/0.019} & \makecell[c]{\textbf{0.021}/\textbf{0.018}, \\ \textbf{0.016}/\emph{\textcolor{blue}{0.015}}}  & \makecell[c]{\textbf{0.025}/\textbf{0.025}, \\ \textbf{0.022}/\textbf{0.020}, \\ \textbf{0.027}/{\textbf{0.024}}} & \makecell[c]{\textbf{0.040}/\textbf{0.019}, \\ \textbf{0.027}/\textbf{0.023}} & \makecell[c]{\textbf{0.005}/\textbf{0.005}, \\ \textbf{0.010}/\textbf{0.009}} & \makecell[c]{\textbf{0.014}/\textbf{0.011}, \\ \textbf{0.023}/\textbf{0.021}} & \makecell[c]{{\textbf{0.035}}/\emph{\textcolor{blue}{0.033}}, \\ \textbf{0.039}/\emph{\textcolor{blue}{0.033}}, \\ \textbf{0.025}/\textbf{0.016}, \\ \textbf{0.013}/\textbf{0.011}}  & \textbf{0.023}/\textbf{0.019}
        \\ \cline{2-10} 
        \specialrule{0em}{1pt}{0pt}

        ~ & Oracle ICP & \makecell[c]{{0.091}/\emph{\textcolor{blue}{0.041}},  \\ {0.070}/\emph{\textcolor{blue}{0.030}}} & \makecell[c]{0.126/\emph{\textcolor{blue}{0.028}}, \\ \emph{\textcolor{blue}{0.032}}/\textbf{0.013}} & \makecell[c]{0.092/0.097, \\ 0.188/0.197, \\ 0.185/0.193} & \makecell[c]{0.071/0.037, \\ 0.120/\emph{\textcolor{blue}{0.030}}} & \makecell[c]{0.072/\emph{\textcolor{blue}{0.036}}, \\ 0.060/\emph{\textcolor{blue}{0.017}}} & \makecell[c]{0.123/0.122,\\ 0.120/0.123} & \makecell[c]{\emph{\textcolor{blue}{0.053}}/\textbf{0.029},\\\emph{\textcolor{blue}{0.054}}/\textbf{0.027},\\\emph{\textcolor{blue}{0.050}}/\emph{\textcolor{blue}{0.028}},\\ \emph{\textcolor{blue}{0.052}}/\emph{\textcolor{blue}{0.031}}}  & 0.092/0.063
        \\ 
        \cline{2-10} 
        \specialrule{0em}{1pt}{0pt}

        ~  & Ours & \makecell[c]{\emph{\textcolor{blue}{0.054}}/0.052, \\ \emph{\textcolor{blue}{0.067}}/0.046} & \makecell[c]{\emph{\textcolor{blue}{0.082}}/0.083, \\ 0.042/0.034}  & \makecell[c]{\emph{\textcolor{blue}{0.054}}/\emph{\textcolor{blue}{0.039}}, \\ \emph{\textcolor{blue}{0.086}}/\emph{\textcolor{blue}{0.088}}, \\ \emph{\textcolor{blue}{0.070}}/\emph{\textcolor{blue}{0.055}}} & \makecell[c]{{\textbf{0.040}}/\emph{\textcolor{blue}{0.037}}, \\ \emph{\textcolor{blue}{0.046}}/0.042} & \makecell[c]{\emph{\textcolor{blue}{0.066}}/0.069, \\ \emph{\textcolor{blue}{0.037}}/0.035} & \makecell[c]{\emph{\textcolor{blue}{0.021}}/\emph{\textcolor{blue}{0.019}}, \\ \emph{\textcolor{blue}{0.027}}/\emph{\textcolor{blue}{0.026}}} & \makecell[c]{{0.096}/{0.096}, \\ {0.097}/{0.092}, \\ {0.108}/{0.105}, \\ 0.109/0.100}  & \emph{\textcolor{blue}{0.065}}/\emph{\textcolor{blue}{0.060}}
        \\ \cline{1-10} 
        \specialrule{0em}{1pt}{0pt}
        
        \multirow{2}{*}{J} & \makecell[c]{NPCS-EPN \\ (supervised)} & \textbf{5.04}/\textbf{0.076} & \textbf{5.66}/\textbf{0.078}  & \makecell[c]{\textbf{7.42}/0.090,\\ \textbf{7.42}/\textbf{0.101}}   & \textbf{5.74}/0.129 & 14.15/0.063 & \textbf{8.53}/\textbf{0.084} & \textbf{20.18}/-  & \textbf{9.27}/\textbf{0.089}
        \\ \cline{2-10}  
        \specialrule{0em}{1pt}{0pt}

        ~ & Ours & 20.30/0.089 & 28.40/0.118  & \makecell[c]{17.75/\textbf{0.045},\\ 17.75/0.129}  & 30.31/\textbf{0.122} & \textbf{4.36}/\textbf{0.031} & 17.17/0.169 & 38.86/-  & 21.86/0.100
        \\ \cline{1-10} 
        \specialrule{0em}{1pt}{0pt}

    \end{tabular}
    }
    \vspace{-16pt}
    \label{tb_exp_pose_cmp}
\end{table}

\subsection{Part Segmentation} \label{sec_exp_seg}
\vpara{Evaluation Metric and Baselines.} 
The metric used for this task is Segmentation IoU (MIoU). 
% We choose two implicit field-based segmentation strategy 
We choose three position-based segmentation strategies, namely BAE-Net~\cite{chen2019bae}, NSD~\cite{kawana2020neural}, BSP-Net~\cite{chen2020bsp} and one motion-based segmentation method ICP~\cite{algo_icp} as our baselines for this task. 
% Baselines of this task are 
% BAE-Net~\cite{chen2019bae}, NSD~\cite{kawana2020neural}, BSP-Net~\cite{chen2020bsp} and ICP~\cite{algo_icp}. 
For BAE-NEt and BSP-Net, we generate data in their implicit representation using the data generation method described in IM-NET~\cite{chen2019learning}. 
% For ICP, the segmentation labelsof each point in the input shape is taken set to the label of their nearest registered part from the example shape directly. 
%%%% Appendix %%%%
We improve the evaluation strategy for NSD and BSP-Net considering the global pose variation of our data (see Appendix~\ref{sec_appen_baselines} for details). 

\vpara{Experimental Results.}
In table~\ref{tb_exp_seg_cmp}, we present experimental results of our method and baselines on complete point clouds. 
Results on partial data are deferred to Table~\ref{tb_exp_seg_cmp_partial} in the Appendix~\ref{sec_appen_exp_partial}. 
Our method can consistently outperform such four part segmentation methods in all categories. 
% It may demonstrate the priority of understanding articulated objects in parts via part motion-related information which is more consistent across shapes with vairous part poses than coordinate-related information. 
BSP-Net, BAE-Net, and NSD assume input data alignment and highly rely on position information for segmentation. 
However, such segmentation cues may not be well preserved in our data with arbitrary global pose variations. 
% For instance, BSP-Net relies on convex index consistency across different shapes to get intra-category aligned part segmentations. 
% rely on the part position consistency for segmentation. Similarly, 
% BSP-Net relies on the consistency of convex indexes across different shapes. Such segmentation cues may not be well preserved when the object pose consistency does not hold on data. 
% One possible reason for BSP's poor performance may come from the inconsistency of convex indexes across different shapes and the intra-category shape variation.
% Besides, we can make the following observations: 
% 1) 
% BSP-Net underperforms our method on all categories.
% % Compared with part segmentation performance of BSP-Net achieved on aligned shapes presented in~\cite{kawana2022uppd}, it degenerates a lot when being tested on shapes with arbitrary global pose variations. 
% One possible reason may come from the inconsistency of convex indexes across different shapes and the intra-category shape variation.
% % Moreover, intra-category shape variation as well as the the change of part articulation states would also result in its degenerated performance on articulated objects. 
By contrast, part motions can serve as a more consistent cue for segmenting articulated objects than positions. 
% considering the nature of our data. 
% for articulated objects than position. 
% -base consistency. 
% than part location.
% as well as the shape variation caused by the change of part articulation states that would make it hard for us to find a proper shape from the training set with consistent segmented convexes. 
% 2) 
We hypothesize that ICP's poor registration performance on some categories such as Eyeglasses further lead to its low segmentation IoUs. 
% The low segmetnation IoU of ICP 
% % The low segmentation IoU of ICP 
% on some categories such as Eyeglasses and Laptop may be related to its poor registeration resutls,
% may be related to its poor
% % not-so-good 
% registeration performance affected by the local minimum in the optimization process and the part symmetry-related problem of articulated objects. 
%%%%% %%%%%
% may be related to
% the unsatisfactory registeration results influenced by local optimum in the optimization process of ICP which is further aggravated by part symmetry. 
%%%%% %%%%%
% Compared to the above two methods, our strategy use part motions as the segmentation cue to perform segmentation which is more consistent across  shapes in the same category and could get better segmentation performance than them. 
% in different articulation states and global pose variation than coordinate-related information. 

% via part motions other than just from geometric information. 
% It can be seen that... \todo{xxx}.
% bounding box 

\subsection{Shape Reconstruction} \label{sec_exp_recon}
\vpara{Evaluation Metric and Baselines.}
We choose to use Chamfer L1 as our evaluation metric for shape reconstruction. 
To demonstrate the superiority of part-by-part reconstruction for articulated objects over the whole shape reconstruction, we choose EPN which treats them as rigid objects for reconstruction as the baseline.
% for this task. 
% , which treats input shapes as rigid objects for shape reconstruction.
%%%% appendix %%%%
% Due to the huge computing resource needed by EPN, especially for extracting per-point features without downsampling the input shape, we use 512 points (480 for Eyeglasses) for each input shape for all methods. 
% % the input shape should be downsampled to a relatively low resolution. 
% % Here, for a fair comparision, the number of points output by the reconstruction branch is set to 512 for each method on all categories, but 480 for Eyeglasses. 
% % The number of points
% For part-by-part reconstruction, the number of points decoded by each decoder is set to the same number of points in the input shape. 
% Then, the reconstruction for the whole shape is obtained by first concatenating all part reconstructions and further downsample to the number of points in the input shape. 
%%%% appendix %%%%
% the whole shape reconstruction is obtained by concatenating all part reconstructions and further downsampled to 512 points. 
% \todo{EPN's heavy computation cost and the reason for setting the number to 512}

\vpara{Experimental Results.}
As shown in Table~\ref{tb_exp_completion_cmp}, our method can consistently outperform the EPN-based whole shape reconstruction. 
We suppose part-by-part reconstruction where only simple parts should be recovered makes the reconstruction an easier problem for networks than recovering the whole shape. 
% is a much more easier problem for networks 
% One possible explanation is that only simple parts are reconstructed in the part-by-part approach, making the reconstruction an easier problem for networks. 
% which turns out to be an easier problem than that of reconstructing the whole shape. 
% which would turn out to be an easier task than reconstructing the whole shape, where the input shape with articulation state variations are needed to be captured. 
% It could also demonstrate the priority of part-based shape understanding for 
% It can demonstrate the priority of part-by-part shape modeling for articulated objects by reconstructing only simple parts other than the whole shape directly, to some extent.
% to some enxtent. 

\begin{table}[h]
    \centering
    \caption{\footnotesize 
    % \textcolor{myblue}{
    Comparison between the part segmentation performance of different methods on all  categories. 
    % }
    Metric used for this task is Segmentation MIoU, calculated on 4096 points for each shape. Values presented in the table are scaled by 100. Larger values indicate better performance. ``*'' denote cases where the network fails by segmenting input shapes into single parts. 
    } 
%   \vspace{-8pt}
  \vspace{-1pt}
	\resizebox{0.7\linewidth}{!}{%
    \begin{tabular}{@{\;}c@{\;}|c|c|c|c|c|c|c|c@{\;}}
    \midrule
        \hline
        \specialrule{0em}{1pt}{0pt} 
        ~ & Oven & \makecell[c]{Washing\\ Machine} & Eyeglasses & Laptop (S) & Safe & Laptop (R) & Drawer & Avg.
        \\ 
        \cline{1-9} 
        \specialrule{0em}{1pt}{0pt}
        
        % \textcolor{myblue}{
        % mean error / Median error
        BAE-Net~\cite{chen2019bae} & 55.04  & 46.07* & 37.19* & 65.21 & 39.83*  & 66.35 & 22.83* & 47.50 \\
        
         % mean error / Median error
        NSD~\cite{kawana2020neural} & 60.59   & 56.43 & 53.31 & 80.88 & 71.30  & 76.86 & 33.61 & 61.85
        \\ 
        % }
        % \cline{1-8} 
        % \specialrule{0em}{1pt}{0pt}
        
        % mean error / Median error
        BSP-Net~\cite{chen2020bsp} & 67.24  & 62.52 & 54.28 & 79.41 & 76.59  & 81.33 & 42.15 & 66.22
        \\ 
        % \cline{1-8} 
        % \specialrule{0em}{1pt}{0pt}

        % mean error / Median error
        Oracle ICP~\cite{algo_icp} & 75.17  & 72.80 & 49.49 & 56.20 & 66.90  & 59.96 & 45.68 & 60.89
        \\ 
        % \cline{1-8} 
        % \specialrule{0em}{1pt}{0pt}

        % mean error / Median error
        Ours & \textbf{76.22}  & \textbf{73.27} & \textbf{62.84} & \textbf{82.97} & \textbf{80.06} & \textbf{86.04} & \textbf{51.39} & \textbf{73.26}
        \\ \cline{1-9} 
        \specialrule{0em}{1pt}{0pt}
        
    \end{tabular}
     }
    %  \vspace{-10pt}
      \vspace{-16pt}
    \label{tb_exp_seg_cmp}
\end{table}

\begin{table}[htbp]
    \centering
    \caption{\footnotesize 
    Comparison between the shape reconstruction performance of different methods on all  categories. Metric used in this task is Chamfer L1. The smaller, the better. 
    } 
  \vspace{-1pt}
	\resizebox{0.8\linewidth}{!}{%
    \begin{tabular}{@{\;}c@{\;}|c|c|c|c|c|c|c|c@{\;}}
    \midrule
        \hline
        \specialrule{0em}{1pt}{0pt} 
        Method & Oven & \makecell[c]{Washing \\ Machine} & Eyeglasses & Laptop (S) & Safe & Laptop (R) & Drawer & Avg. \\ 
        \cline{1-9} 
        \specialrule{0em}{1pt}{0pt}
        
        % mean error / Median error
        EPN~\cite{li2021leveraging} & {0.033} & 0.051 & {0.028} & {0.029}  & 0.030 & {0.028} & 0.057 & 0.036
        \\ 
        % \cline{1-8} 
        % \specialrule{0em}{1pt}{0pt}
        
        Ours & \textbf{0.025} & \textbf{0.049} & \textbf{0.025} & \textbf{0.024} & \textbf{0.026} & \textbf{0.026} & \textbf{0.045} & \textbf{0.031}
        \\ \cline{1-9} 
        \specialrule{0em}{1pt}{0pt}
      
    \end{tabular}
    }
    % \vspace{-10pt}
    \vspace{-16pt}
    \label{tb_exp_completion_cmp}
\end{table}

\begin{table}[htbp]
  \centering
  \caption{\footnotesize 
  Ablation study w.r.t. the effectiveness of joint reguralization for part pose estimation and the design of pose-aware equivariant feature communication (denoted as ``Pose.''). Reported values are per-category per-part average values. 
%   }
  Please refer to the caption of Table~\ref{tb_exp_abl_with_part_proposal} for the data format of ``Joint''.
  }  
%   \vspace{-8pt}
\resizebox{0.8\linewidth}{!}{%
  \begin{tabular}{@{\;}c@{\;}|c|c|c|c|c|c|c@{\;}}
  \midrule
      \hline
      \specialrule{0em}{1pt}{0pt} 
      Method & Seg. IoU & Mean $R_{err}(^\circ)$ & Median $R_{err}(^\circ)$ & Mean $T_{err}$ & Median $T_{err}$ & Joint & Chamfer L1 
      \\ 
      \cline{1-8} 
      \specialrule{0em}{1pt}{0pt}
   
      % mean error / Median error
      No $\mathcal{L}_{reg}$ & {76.40}  & {11.74} & 10.87 &  0.070 & 0.065 & - & 0.038
      \\ 
    %   \cline{1-8} 
    %   \specialrule{0em}{1pt}{0pt}
      
      With $\mathcal{L}_{reg}$  &  74.32   & 10.40 & 9.30 &  0.072 & 0.073 & 22.01/0.111 & 0.032
      \\ 
    %   \cline{1-8} 
    %   \specialrule{0em}{1pt}{0pt}
      
      With $\mathcal{L}_{reg}$ (Pose.)  & \textbf{76.90}  & \textbf{9.21} & \textbf{8.40} & \textbf{0.052} &  \textbf{0.047} & \textbf{19.72/0.103} & \textbf{0.025}
      \\ \cline{1-8} 
      \specialrule{0em}{1pt}{0pt}
      
  \end{tabular}
  }
  \vspace{-18pt}
  \label{tb_exp_abl_iteration_joint}
\end{table}

\begin{table}[htbp]
  \centering
  \caption{\footnotesize 
  Ablation study w.r.t. the effectiveness of accumulating part-level features for part-based properties prediction. Reported values are per-category per-part average values on all categories. 
%   }
  ``Joint'' represents joint parameter estimation errors, with the value in the format of 
%   Metrics presented for joint prediction are in the format of 
  ``Mean $\theta_{err}$/Mean $d_{err}$''. 
  } 
  \vspace{-2pt}
\resizebox{0.8\linewidth}{!}{%
  \begin{tabular}{@{\;}c@{\;}|c|c|c|c|c|c|c@{\;}}
  \midrule
      \hline
      \specialrule{0em}{1pt}{0pt} 
      Method & Seg. IoU & Mean $R_{err}(^\circ)$ & Median $R_{err}(^\circ)$ & Mean $T_{err}$ & Median $T_{err}$ & Joint & Chamfer L1
      \\ 
      \cline{1-8} 
      \specialrule{0em}{1pt}{0pt}

      Without Parts &  71.68  & 12.85 & 11.52 & 0.068 & 0.060 & {27.97/0.172}   & 0.036 \\

      With Parts & \textbf{76.90}  & {\textbf{9.21}} & \textbf{8.40}  & \textbf{0.052} & \textbf{0.047} & \textbf{19.72/0.103} & \textbf{0.029}
    %   \\ \cline{1-7} 
    %   \specialrule{0em}{1pt}{0pt}
      \\ \cline{1-8} 
      \specialrule{0em}{1pt}{0pt}
    
  \end{tabular}
  }
  \vspace{-18pt}
  \label{tb_exp_abl_with_part_proposal}
\end{table}

% !TEX root = ../main.tex

\section{Ablation Study}
% \begin{itemize}
%   \item Iterative pose estimation
%   \item The effectiveness of equivariance --- (global align + KPConv) --- especially for eyeglasses?
%   \item Joint meodeling ---> cmp between with joint modeling and without joint modeling --> especially for translation error ---> part movement
%   \item with part proposal ---> use features from the whole shape for calculation.

In this section, we try to ablate some crucial designs in the method to demonstrate their effectiveness, 
including part-level feature accumulation, pose-aware point convolution, and joint regularization.

% \vpara{Part-level Features.}
\vpara{Part-level Feature Accumulation.}
We use a grouping module to group points into parts for part-level features in our method. 
% Then, the part-level features can be obtained by grouping per-point features of points assigned to the part. 
To demonstrate the effectiveness of using part-level features for part shape, structure, pose disentanglement,
% % Then, for each part, we group features of points assigned to the part for its part-level features. 
% We expect that part-level features can help with disentanglement between part pose and part shape, thus facilitating further shape reconstruction and pose estimation. Therefore,
%%%%% %%%%%
% we ablate part-level features in the design by just using features from the whole shape for part properties prediction, like the design in~\cite{kawana2022uppd,chen2019bae}.
%%%%% %%%%%
we ablate part-level features and only use features from the whole shape for part-level properties prediction, similar to those used in~\cite{kawana2022uppd,chen2019bae}. 
Table~\ref{tb_exp_abl_with_part_proposal} compares their performance.
% between 
% % summarizes the performance comparison between 
% the method using part-level features and the one using features from the whole shape on tested categories. 
For each metric, we report its per-category per-part average value. 
% on the Laptop (R).
% It can be observed
It can be observed that part-level features can help with part-based properties prediction, letting the network achieve better performance on all pose-related metrics.

\vpara{Pose-aware Point Convolution.}
Our method contains a pose-aware equivariant feature convolution design for part-level SE(3) equivariant feature learning. 
% In the proposed method, we add a pose-aware equivariant feature communication strategy to design our part-level equivariant network based on EPN.
% to seek for ``part-level'' equivariance. 
% to estimate part pose iteratively for better equivariant features with the pose change of contextual parts disentangled.
%  feature extraction and further part-level properties. 
% The motivation is to 
% todo: add some motivations?
To demonstrate the superiority of part-level equivariance over common global equivariance,
% using part-level equivariant features to solve the articulated object pose estimation problem,
% the part pose estimation problem, 
we compare the model's performance when using
% such features 
part-level equivariant features
(With $\mathcal{L}_{reg}$ (Pose.)) with the one using global equivariant features (With $\mathcal{L}_{reg}$) in table~\ref{tb_exp_abl_iteration_joint}. 
For each metric, its per-category per-part average value is reported. 
% we report its per-category per-part average values. 
% on Laptop (R). 
% In table~\ref{tb_exp_abl_iteration_joint}, we compare the performance of the proposed strategy with/without the pose-aware inter-part feature communication design on two categories, 
% iterative pose estimation design on two categories, 
% namely Oven and Laptop (R).
% It can be seen that the performance of the method on segmentation task and pose estimation task get better on most of tested metrics. 
% Performance improvement on segmentation IoU and most pose-related metrics can then prove our assumption. 
The network using part-level equivariant features could consistently outperform the one using only global equivariant features on all metrics. 

% \subsection{Equivariant Network for Part Pose Estimation}
% \todo{EPN + KPConv?} --> perhaps add this part to the appendix?

% \vpara{Reguralization for Joint Prediction.}
\vpara{Joint Regularization.}
% To predict the rigid transformation of revolute parts and prismatic parts, we propose to model their articulated transformation as rotating around a joint and transforming along an axis respectively. 
% To predict part rigid transformation, we propose to model their articulation transformation based on predicted joints.
% their rigid motions as rotating around a joint and transforming along an axis, respectively, and then predict parameters for such properties to further calculate part rigid transformation. 
% In the method, we predict part pose by predicting 
% equivariant and/or invariant 
% joint parameters as well as part motion parameters along the joint such as rotation angles. 
% To predict the part rigid transformations in the canonical object space, we first predict joint parameters and then rigid transformation based on predicted joints. 
Besides reconstruction loss, we additionally add a joint regularization term to predict joints that connect two adjacent parts. 
% Further, we add a regularization term to make the predicted joints behave like ``real'' joints (see sec.~\ref{sec_method_self_supervised_task}). 
% Further, to make the predicted joint behave like a ``real'' joint passing through two parts, we add 
% However, the predicted joint may not perform like a ``real'' joint passing through two parts if we just use the predicted parameters, like axis direction, angle, pivot point to calculate part rigid transformations. 
% Therefore, we propose to 
% a reguralization term for the predicted joint based on the assumption that a joint should pass through two parts it connects together. 
% It seems that it is not a necessary design if we only aim at estimating category-level part 6D pose, but a requirement to recover the real part kinematic chain. 
% However, we do observe better part pose estimation results predicted by the joint-aware strategy, compared with those without joint modeling. 
Beyond acquiring joint-related parameters, joint regularization could improve the pose estimation performance as well, especially for translation prediction, as shown in Table~\ref{tb_exp_abl_iteration_joint}. 
% we also observe the performance improvement on pose estimation with joint regularization, especially on translation estimation error, as shown in Table~\ref{tb_exp_abl_iteration_joint}. 
%%%%% %%%%%
% Besides obtaining joint parameter prediction, we observe that the joint reguralization can also help improve the pose estimation results. 
% In table~\ref{tb_exp_abl_iteration_joint}, we compare the experimental results of the method with joint reguralization and without joint reguralization. 
% When predicting joints for 
% calculating part rigid transformations,
% % part transformation calculation, 
% we can arrive at results consistently better than the strategy using no joint reguralization, especially on part translation estimation error. 
%%%%% %%%%%
% One possible reason is the relationship between two parts' translations and rotations bridged by the estimated joint can help better part pose estimation. 

% !TEX root = ../main.tex

% \section{Conclusion and Limitations} 
\section{Conclusion} 
\label{sec_conclusion}
In this work, we propose a self-supervised strategy for category-level articulated object pose estimation without any annotations.
% any access to CAD models, part segmentation labels, or pose-related parameters. 
Leveraging part-level SE(3) equivariant features, we propose a part shape, structure, pose disentanglement strategy that successfully 
% decompose such information for 
accomplish the category-level articulated object pose estimation task. 
A part-by-part shape reconstruction task is adopted to self-supervise the network learning.
% part-by-part shape reconstruction task to self-supervise the network to do the disentanglement of part shape, structure, and pose from input shapes. 
% By using SE(3)-invariance, the network can then reconstruct a reference frame for each part in its own canonical part space, and a reference frame for the whole shape in the canonical object space. 
% Together with SE(3)-equivariance for part pose estimation, we can solve the problem of articulated object pose estimation in an unsupervised manner.
Experiments prove the effectiveness of our method and our core ideas. 
% our proposed self-supervised strategy and the benefit of using part-level equivariance for this problem. 
% articulated object pose estimation. 
% By using SE(3)-equivariance and invariance, we let the network reconstruct a reference frame for each part in its canonical space.
% Then, we predict the rigid transformation to transform each part from its canonical space to the camera space to complete the shape reconstruction-based task to self-supervise the network's learning process.
% Moreover, we propose a pose-aware inter-part equivariant feature communication strategy for part-level equivariant features. 
This work can reduce the annotation efforts for solving 
this tasks and would also promote further thinkings on designing part-level equivariant networks. 

\bibliographystyle{iclr2023_conference}
\bibliography{ref}

\clearpage

\appendix

% !TEX root = ../main.tex

The appendix is organized as follows:
\begin{itemize}[noitemsep,topsep=0pt,parsep=0pt,partopsep=0pt,leftmargin=.5cm]
    \item Proofs and further explanations on the method
    \begin{itemize}[noitemsep,topsep=0pt,parsep=0pt,partopsep=0pt,leftmargin=.5cm]
        \item 
        % Proof of part-level equivariant design
        Proof of part-Level equivariant property of the pose-aware point convolution module (sec.~\ref{sec_appen_part_level_equiv}).
        \item Further explanations on some method components (sec.~\ref{sec_appen_additional_explanations}).
    \end{itemize}
    \item Experiments
    \begin{itemize}[noitemsep,topsep=0pt,parsep=0pt,partopsep=0pt,leftmargin=.5cm]
        \item Data preparation (sec.~\ref{sec_appen_data_prepare}).
        \item Implementation details (sec.~\ref{sec_appen_imple_details}).
        \item Implementation details for baselines (sec.~\ref{sec_appen_baselines}).
        \item Experiments on partial point clouds (sec.~\ref{sec_appen_exp_partial}).
        \item Additional comparisons and applications (sec.~\ref{sec_appen_more_exp_additional}).  % experimental comparisons
        \item Robustness to input data noise (sec.~\ref{sec_appen_robustness}).
        \item Visualization of part-level equivariant features (sec.~\ref{sec_appen_part_level}).
        \item Evaluation strategy for category-level articulated object poses (sec.~\ref{sec_appen_eval}).
    \end{itemize}
    \item Discussion on part symmetry (sec.~\ref{sec_appen_symmetric_parts})
\end{itemize}

\section{Method Details}

\begin{figure*}[ht]
    \centering
      \includegraphics[width=0.95\textwidth]{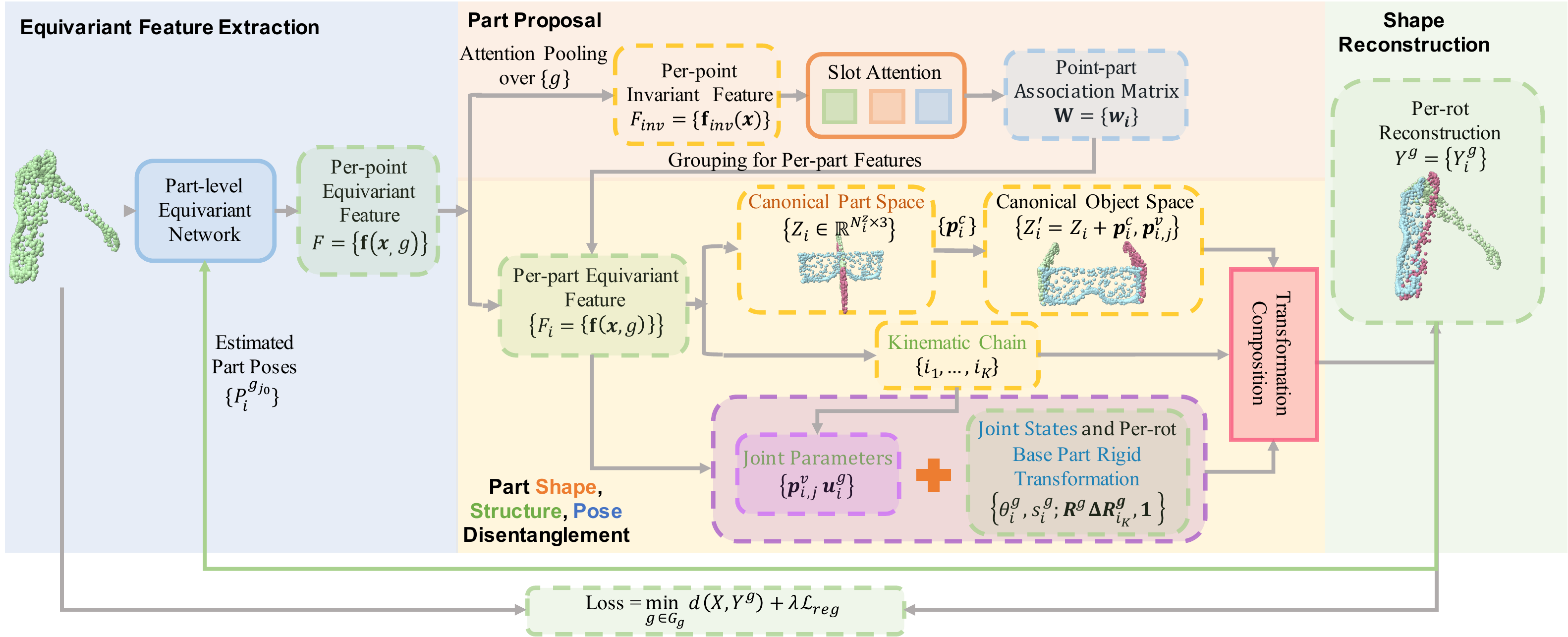}
      \vspace{-10pt}
    \caption{
      \footnotesize
    %   \textcolor{}{}
    % \textcolor{myblue}{
      Overview of the proposed self-supervised articulated object pose estimation strategy.
    The method takes a complete or partial point cloud of an articulated object as input, factorizes canonical shapes, object structure, and the articulated object pose from it. 
    The network is trained by a shape reconstruction task. 
    Part-level SE(3) equivariant features are learned  by iterating between part pose estimation and pose-aware equivariant point convolution. 
      \textcolor{mygreen}{Green} lines (\textcolor{mygreen}{$\leftarrow$}) denote procedures for feeding the estimated part poses back to the  pose-aware point convolution module.
    }
    \label{fig_overall_pipeline_detailed}
    \vspace{-10pt}
\end{figure*}

\subsection{Overview} \label{sec_appen_method_detailed_overview}
We provide an detailed diagram of our self-supervised learning strategy in Figure~\ref{fig_overall_pipeline_detailed}.

\subsection{Proof of Part-Level Equivariant Property of the Pose-aware Point Convolution Module} \label{sec_appen_part_level_equiv}
% \vpara{``Part-level'' Equivariant Features.}
% In the method, we propose to use the estimated part pose and predicted part segmentation into the feature convolution process for better ``part-level'' equivariant features. 
% Features of such property should change equivalent to the pose change of itself, but invariant to the pose change of its contextual parts. 
% Since the group convolution of EPN~\cite{chen2021equivariant} does not communicate features between different points, we propose to modify the point convolution of EPN using the estimated part poses.
In this section, we prove the part-level equivariant property of the designed pose-aware point convolution module:
% Formally, 
% \begin{align}
%   (\mathcal{F} * h_1)(x,g) = \sum_{\mathcal{P}_i^{-1}x_i\in \mathcal{N}_{\mathcal{P}^{-1}x}^c} \mathcal{F}(x_i, g \mathcal{R}\mathcal{R}_i^{-1} ) h_1(g(x - \mathcal{P}\mathcal{P}_i^{-1}x_i)),
% \end{align}
\begin{align}
  (\mathcal{F} * h_1)(x_i,g) = \sum_{{P}_j^{-1}x_j\in \mathcal{N}_{{P}_i^{-1}{x_i}}^c} \mathcal{F}(x_j, g \mathbf{R}_i\mathbf{R}_j^{-1} ) h_1(g(x_i - {P}_i{P}_j^{-1}x_j)),
\end{align}
where ${P}_i$ and ${P}_j$ are the (estimated) pose of $x_i$ and $x_j$ from the canonical object space to the camera space respectively, $\mathbf{R}_i$ and $\mathbf{R}_j$ are the (estimated) rotations of point $x_i$ and point $x_j$ from the canonical object space to the camera space respectively, $\mathcal{N}_{{P}_i^{-1}x_i}^c$ denotes the set of point $x_i$'s neighbours in the canonical object space.
Note that the neighbourhood set $\mathcal{N}_{x_i}$ in the Equation~\ref{eq_part_level_convolution} 
% refers to $x_i$'s neighbourhood in the camera space, which 
represents the neighbourhood of $x_i$ in the camera space with points in which belong to $x_i$’s neighourhood in the canonical object space, \emph{i.e.} $\mathcal{N}_{{P}_i^{-1}{x_i}}^c$. 
$\mathcal{N}_{x_i}$ would vary as $x_i$'s pose changes, while $\mathcal{N}_{{P}_i^{-1}{x_i}}^c$ keeps the same. 
$\{ x_j \vert x_j \in \mathcal{N}_{x_i} \} = \{ x_j \vert {P}_j^{-1}x_j\in \mathcal{N}_{{P}_i^{-1}{x_i}}^c \}$. 
% change with $x_i$'s pose. 
% The point set $\mathcal{N}_{{P}_i^{-1}{x_i}}^c$ denotes $x_i$'s neighbourhood in the canonical object space and would not change to $x_i$'s pose variations. 
% In Equation~\ref{eq_part_level_convolution}, $\mathcal{N}$
% which is ${P}_j^{-1}x_j\in \mathcal{N}_{{P}_i^{-1}{x_i}}^c$ actually as we use in the above equation. 

% inversedly transformed x_i and others...
To prove the part-level equivariance of $(\mathcal{F}* h_1)(x_i,g)$, we need to prove 1) $(\mathcal{F}* h_1)(x_i,g)$ is invariant to the rigid transformation 
% to the pose change 
of point $x_i$'s each neighbouring point $x_j$;
% i.e. $\mathcal{P}_i = (\mathcal{R}_i, \mathcal{T}_i)$; 
2) $(\mathcal{F}* h_1)(x_i,g)$ is equivariant to the rigid transformation of $x_i$ itself. 
% i.e. $\mathcal{F} * h_$

We then prove those properties for the continuous convolution operations, 
$(\mathcal{F} * h_1)(x_i, g) = \int_{x_j\in \mathbb{R}^{3}} \mathcal{F}(x_j, g\mathbf{R}_i\mathbf{R}_j^{-1})h_1(g(x_i - {P}_i{P}_j^{-1}x_j))$. 

\begin{theorem}
    The continuous operation $(\mathcal{F} * h_1)(x_i, g) = \int_{x_j\in \mathbb{R}^{3}} \mathcal{F}(x_j, g\mathbf{R}_i\mathbf{R}_j^{-1})h_1(g(x_i - {P}_i{P}_j^{-1}x_j))$ is invariant to each arbitrary rigid transformation 
    % to the rigid transformation 
    $\Delta {P}_j = (\Delta \mathbf{R}_j \in \text{SO(3)}, \Delta \mathbf{t}_j \in \mathbb{R}^{3})$ of ($x_j \forall x_j\in \mathbb{R}^3, x_j\neq x_i$)
    % applied on each 
    of $x$'s neighbouring point $x_j$.
\end{theorem}

\begin{proof}
To prove the invariance of $(\mathcal{F} * h_1)(x_i, g)$, we need to prove that $\forall x_j \in \mathbb{R}^{3}, x_j\neq x_i, \forall \Delta{P}_j \in \text{SE(3)}, \mathbf{R}_j' = \Delta \mathbf{R}_j \mathbf{R}_j$, we have 
\begin{align*}
    \Delta {P}_j (\mathcal{F} * h_1) (x_i,g) = (\mathcal{F} * h_1) (x_i,g).
    % (\mathcal{F} * h_1)(x,g) = \int_{x_i\in \mathbb{R}^{3}} \int_{g_j\in \text{SO(3)}} \mathcal{F}(\Delta \mathcal{P}_i x_i, g\mathcal{R}\mathcal{R}_i^{-1}\Delta \mathcal{R}_i^{-1}) h_1(g(x - \mathcal{P}\mathcal{P}_i^{-1}\Delta \mathcal{P}_i^{-1}x_i)),
\end{align*}
Let $x_j' = \Delta {P}_j x_j$, ${P}_j' = \Delta {P}_j {P}_j$,
% where $\Delta \mathcal{P}_i$ is the change of pose of the neighbouring point $x_i$, 
then we have,
\begin{align*}
    % (\Delta\mathcal{P}_i)_{x_i\in \mathbb{R}^{3}}
    \Delta {P}_j
    (\mathcal{F} * h_1)(x_i, g) &= \int_{x_j'\in \mathbb{R}^{3}}  \mathcal{F}(x_j', g\mathbf{R}_i\mathbf{R}_j^{'-1}) h_1(g(x_i - {P}_i{P}_j^{'-1}x_j')) \\ 
    &=
    \int_{x_j\in \mathbb{R}^{3}}  \mathcal{F}(\Delta {P}_j x_j, g\mathbf{R}_i\mathbf{R}_j^{-1}\Delta \mathbf{R}_j^{-1}) h_1(g(x_i - {P}_i{P}_j^{-1}\Delta {P}_j^{-1}\Delta {P}_jx_j)) \\ 
    &= 
    \int_{x_j\in \mathbb{R}^{3}}  \mathcal{F}(\Delta \mathbf{R}_j x_j, g\mathbf{R}_i\mathbf{R}_j^{-1}\Delta \mathbf{R}_j^{-1}) h_1(g(x_i - {P}_i{P}_j^{-1}x_j))  \\
    &= 
    \int_{x_j\in \mathbb{R}^{3}}  \mathcal{F}(x_j, g\mathbf{R}_i\mathbf{R}_j^{-1}) h_1(g(x_i - {P}_i{P}_j^{-1}x_j)) \\
    &=
    (\mathcal{F} * h_1)(x_i,g).
\end{align*}
% To prove that $\mathca$
\end{proof}

\begin{theorem}
    The continuous operation $(\mathcal{F} * h_1)(x_i, g) = \int_{x_j\in \mathbb{R}^{3}} \mathcal{F}(x_j, g\mathbf{R}_i\mathbf{R}_j^{-1})h_1(g(x_i - {P}_i{P}_j^{-1}x_j))$ is equivariant to the rigid transformation $\Delta {P}_i = (\Delta \mathbf{R}_i \in \text{SO(3)}, \Delta \mathbf{t}_i \in \mathbb{R}^{3})$ 
    of $x_i$.
\end{theorem}

\begin{proof}
To prove that $(\mathcal{F} * h_1)(x_i,g)$ is equivariant to the rigid transformation of $x_i$, we need to prove that $\forall \Delta {P}_i\in \text{SE(3)}$, we have 
\begin{align*}
    \Delta {P}_i (\mathcal{F} * h_1)(x_i,g) =  (\Delta \mathbf{R}_i\mathcal{F} * h_1)(x_i, g).
\end{align*}
It can be proved by
    \begin{align*}
        \Delta {P}_i (\mathcal{F} * h_1)(x_i,g) &=  (\mathcal{F} * h_1)(\Delta{P}_ix_i, g\Delta\mathbf{R}_i) \\ 
        &=  \int_{x_j\in \mathbb{R}^{3}}  \mathcal{F}(x_j, g\Delta \mathbf{R}_i \mathbf{R}_i\mathbf{R}_j^{-1}) h_1(g(\Delta{P}_ix_i - \Delta{P}_i{P}_j^{-1}x_j)) \\ 
        &=  \int_{x_j\in \mathbb{R}^{3}}  \mathcal{F}(x_j, (g\Delta \mathbf{R}_i) \mathbf{R}_i\mathbf{R}_j^{-1}) h_1((g\Delta\mathbf{R}_i)(x_i - {P}_j^{-1}x_j)) \\
        &= (\Delta \mathbf{R}_i\mathcal{F} * h_1)(x_i, g).
    \end{align*}
\end{proof}

\subsection{Further Explanations on some Method Components} \label{sec_appen_additional_explanations}

\begin{figure*}[ht]
    \centering
      \includegraphics[width=0.9\textwidth]{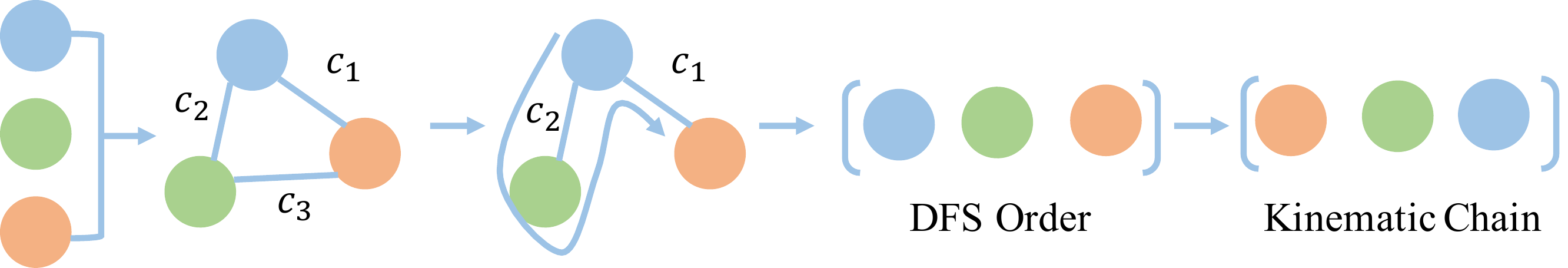}
      \vspace{-10pt}
    \caption{
      \footnotesize
      Kinematic chain prediction procedure (an example of the object containing three parts). 
    }
    \label{fig_kinematic_chain_prediction}
    % \vspace{-5pt}
    \vspace{-10pt}
\end{figure*}

\vpara{Kinematic Chain Prediction.}
The kinematic chain is predict as an invariant property from per-part invariant features to describe part articulation transformation order. 
It is predicted through the following four steps: 
1) Predict an adjacency confidence value $c_{i,j}$ for each part pair $(i,j)$; 
2) Construct an fully-connected adjacency confidence graph $\mathcal{G} = (\mathcal{V}, \mathcal{E})$ based on predicted confidence values, with all parts as its nodes and predicted confidence values as edge weights; 
3) Find a maximum spanning tree from the constructed graph $\mathcal{G}$: $\mathcal{T} = (\mathcal{V}, \mathcal{E}_{\mathcal{T}})$; 
4) Calculate the DFS visiting order of $\mathcal{T}$ and take the inverse visiting order as the predicted kinematic chain. 
We draw the prediction procedure in Figure~\ref{fig_kinematic_chain_prediction}. 

\vpara{Invariant/Equivariant Features for Prediction.}
Given per-part equivariant feature output from the feature backbone $F_i$, its equivariant feature for equivariant properties prediction is further calculated from an SO(3)-PointNet, \emph{i.e.} $\hat{F}_i = \text{SO(3)-PointNet}(X_i, F_i)$. 
Its invariant feature for invariant properties prediction is then computed through a 
% could be obtained through a 
max-pooling operation: $F_i^{inv} = \text{Max-Pooling}(\hat{F}_i)$.

\vpara{Joint Axis Orientation.}
We assume that all joints' axis orientations 
in one shape are consistent. 
Thus, in practice, we set the
orientation of all joints to the same predicted orientation,
% a same joint direction,
\emph{i.e.}
$\mathbf{u}_i^{g}\leftarrow \mathbf{u}_{i_m}^g, \forall (i, j)\in \mathcal{E}_{\mathcal{T}}, \forall g\in G_g$, where $(i_m, j_m)$ is set to the part pair connected to the tree root. 
% to the connected pair with the tree root as one of its node.
% near the tree root.  
Saying $(i,j)\in \mathcal{E}_{\mathcal{T}}$, we mean a directional edge from part $i$ to part $j$. 
In the node pair $(i,j) \in \mathcal{E}_{\mathcal{T}}$, node $i$ is deeper than node $j$ in the tree $\mathcal{T}$.
% where the node $i$ is deeper than node $j$ in the tree $\mathcal{T}$.
It indicates that node $i$'s subtree should rotate around the joint $\mathbf{u}_i^g$ passing through the joint between $i$ and $j$.
% node $j$. 
% rotates around the joint $\mathbf{u}_i^g$ passing through node $j$. 

% \vpara{Prismatic Parts.}
% For shapes having only prismatic parts, the part proposal module is not used.
% Instead, we feed features of the whole shape to the following modules directly to predict properties for each part.
% Thus, to get part segmentations for shapes with prismatic parts like drawers, we adopt a reconstruction-based part label assignment strategy by directly assigning the segmentation label for each point $\mathbf{x}$ in the original shape $X$ as the label of its nearest reconstructed part. 

\vpara{Iterative Pose Estimation.}
Our pose-aware equivariant point convolution module requires per-point pose as input. 
Due to our self-supervised setting where input poses are not assumed, we adopt an iterative pose estimation strategy.
Through this design, we can improve the quality of part-level equivariant features gradually by feeding back estimated poses in the last iteration to the pose-aware point convolution module in the current iteration. 
% which gradually improve the quality of part-level equivariant features by feeding back estimated poses in the previous loop to the pose-aware point convolution module. 
It is because that more accurate input per-point poses would lead to better ``part-level'' SE(3) equivariant features considering the nature of our pose-aware point convolution. 
In practice, we set per-point poses to identity values in the first iteration due to the lack of estimated poses, \emph{i.e.} $P_0 = (\mathbf{R}_0, \mathbf{t}_0)$.  
% per-point poses are set to identity in the first iteration, \emph{i.e.} $P_0 = (\mathbf{R}_0, \mathbf{t}_0)$.  
% In practice, an additional zero iteration is added by treating the whole shape as a single rigid part and estimating its pose $P_0 = (\mathbf{R}_0, \mathbf{t}_0)$.
% Then its inverse rigid transformation $P_0^{-1}$ is 
% used to transform the input shape for the following part pose estimation iterations. 

\begin{figure*}[ht]
    \centering
      \includegraphics[width=0.7\textwidth]{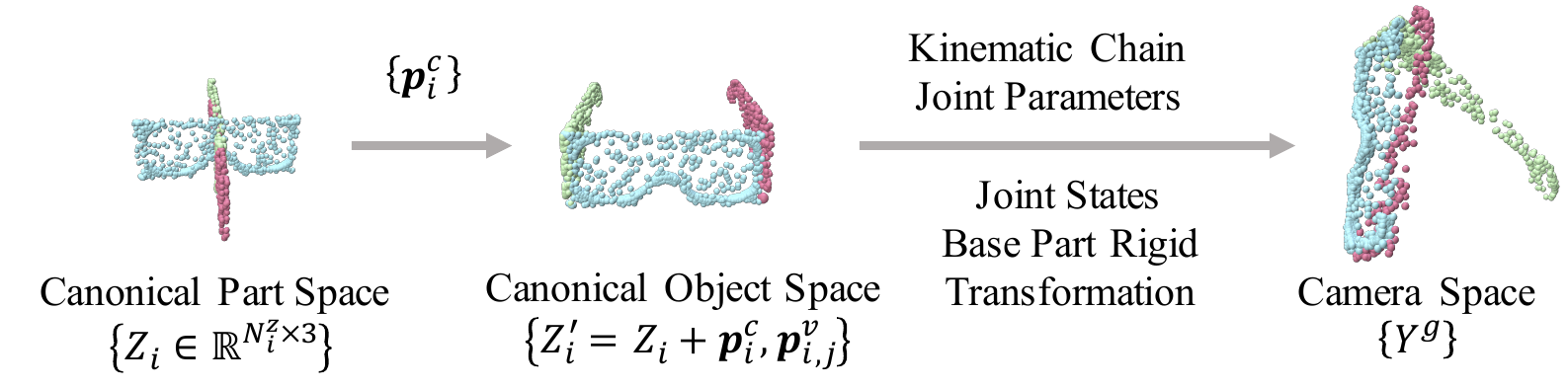}
    %   \vspace{-10pt}
    \caption{
      \footnotesize
      The relationship between our three crucial spaces: the canonical part spaces, the canonical object space, and the camera space. 
    }
    \label{fig_three_spaces}
    % \vspace{-5pt}
    % \vspace{-10pt}s
\end{figure*}

\vpara{Canonical Part Spaces, Canonical Object Space, Camera Space.} 
For each part, the canonical part space normalizes its pose. 
For each object, the canonical object space normalizes its object orientation, articulation states. 
Camera space denotes the observation space. 
Each part shape in the canonical part space is its canonical part shape. 
Each object shape with articulation states canonicalized is called its canonical object shape. 
Canonical \emph{spaces} are category-level concepts, while canonical shapes are instance level concepts. 
Figure~\ref{fig_three_spaces} draws the relationship between such three spaces mentioned frequently in our method.

\vpara{Partial Point Clouds.}
The loss function used in~\cite{li2021leveraging} for partial point clouds is the unidirectional Chamfer Distance. 
Using this function can make the network aware of complete shapes by observing partial point clouds from different viewpoints.
Such expectation can be achieved for asymmetric objects if the viewpoint could cover the full SO(3) space. 
However, we restrict the range of the viewpoint when rendering partial point clouds for articulated objects to make each part visible.
Such restriction would result in relatively homogeneous occlusion patterns. 
% Therefore, we choose to use undirectional Chamfer Distance for partial point clouds, as well.
% The restriction added to the range of view points may cause occlusions patterns that are not that diverse. 
Therefore, we choose to use unidirectional Chamfer Distance only for certain categories such as Safe when tested on partial point clouds. 

\vpara{Equivariant/Invariant Properties of the Designed Modules.}
% \todo{xxx}
The designed method wishes to use part-level SE(3)-equivariance to reduce the difficulty of factorizing part pose and part shape. 
Exact part-level equivariant features can make those modules 
meet our expectations.
However, due to the approximate SE(3)-equivariance of the employed feature backbone and the estimated part pose that may not be very accurate, we cannot expect such invariance/equivariance for them.
For instance, if we do not consider part kinematic constraints, the part shape reconstruction module and the part-assembling parameters prediction module should be invariant to K rigid transformations in the quotient group $(\text{SE(3)}/G_A)(\text{SE(3)})^{K-1}$ if using global equivariant features, while it should be invariant to the rigid transformation in the quotient group $\text{SE(3)}/G_A$ if using part-level equivariant features given correct part pose estimation.
Similarly, the pivot point prediction module should be invariant to two rigid transformations in the quotient group $(\text{SE(3)} / G_A)^2$ if using part-level equivariant features.
Part-level equivariance design could reduce the difficulty of a network doing factorization, which may count as a reason  for its effectiveness. 

\section{Experiments}

\subsection{Data Preparation} \label{sec_appen_data_prepare}
%%%% data collection %%%%
\vpara{Data Collection.} We choose seven categories from three different datasets, namely Oven, Washing Machine, Eyeglasses, Laptop (S) with revolute parts from Shape2Motion~\cite{wang2019shape2motion}, Drawer with prismatic parts from SAPIEN~\cite{xiang2020sapien}, Safe and Laptop (R) with revolute parts from HOI4D~\cite{liu2022hoi4d}. 

The first five datasets are selected according to previous works on articulated object pose estimation or part decomposition~\cite{li2020category,kawana2022uppd}. 
To further test the effectiveness of our method on objects collected from the real world, we choose two more categories (Safe and Laptop (R)) from a real dataset~\cite{liu2022hoi4d}. 

\vpara{Data Splitting.} We split our data according to the per-category data split approach introduced in~\cite{li2020category}. 
Note that not all shapes in a category are used for training/testing. 
Incomplete shapes or instances whose canonical articulation states are inconsistent with other shapes are excluded from experiments. 
% due to their canonical articulation states that are inconsistent with other shapes or missing parts. 
Per-category train/test splits are listed in Table~\ref{tb_exp_dataset_meta_info}.
% For each category, the number of shapes in the train/test sets are listed in 
% % Number of shapes in the train/test sets are listed in
% Table~\ref{tb_exp_dataset_meta_info}.

\vpara{Data Preprocessing.} For each shape, we generate 100 posed shapes in different articulation states. 

Then for complete point clouds, we randomly generate 10 rotated samples from each articulated posed object.
% with each of them rotated by a random rotation matrix. 
% To avoid the ambiguity caused by shape's global symmetry, 
% To avoid global symmetry-related ambiguity for several categories, 
% % such as a almost closed oven may be treated as a cuboid resulting in the ambiguity in the estimated pose, 
% we would restrict the articulation 
When generating articulated posed objects, we would add restrictions on valid articulation state ranges. 
% Specifically, 
For Oven, Safe, and Washing Machine, the valid degree range of their lids is [45$^\circ$, 135$^\circ$).
For Eyeglasses, the range of the degree between two legs and the frame is set to [0$^\circ$, 81$^\circ$). 
For Laptop (S) and Laptop (R), the range of the degree between two parts is set to 
% (9$^\circ$, 99$^\circ$). 
[9$^\circ$, 99$^\circ$). 
% The dataset will be made public. 
% For Safe, the range of the degree betweent the lid and body is set to 

For partial point clouds, we render depth images of complete object instances using the same rendering method described in~\cite{li2021leveraging}. 
The difference is that we manually set a viewpoint range for each category to ensure that all parts are visible in the rendered depth images. 
For each articulated posed shape, we render 10 depth images for it.
% , the same number for generating shapes with arbitrary global poses. 
The dataset will be made public. 

\begin{figure*}[ht]
  \centering
    \includegraphics[width=1.0\textwidth]{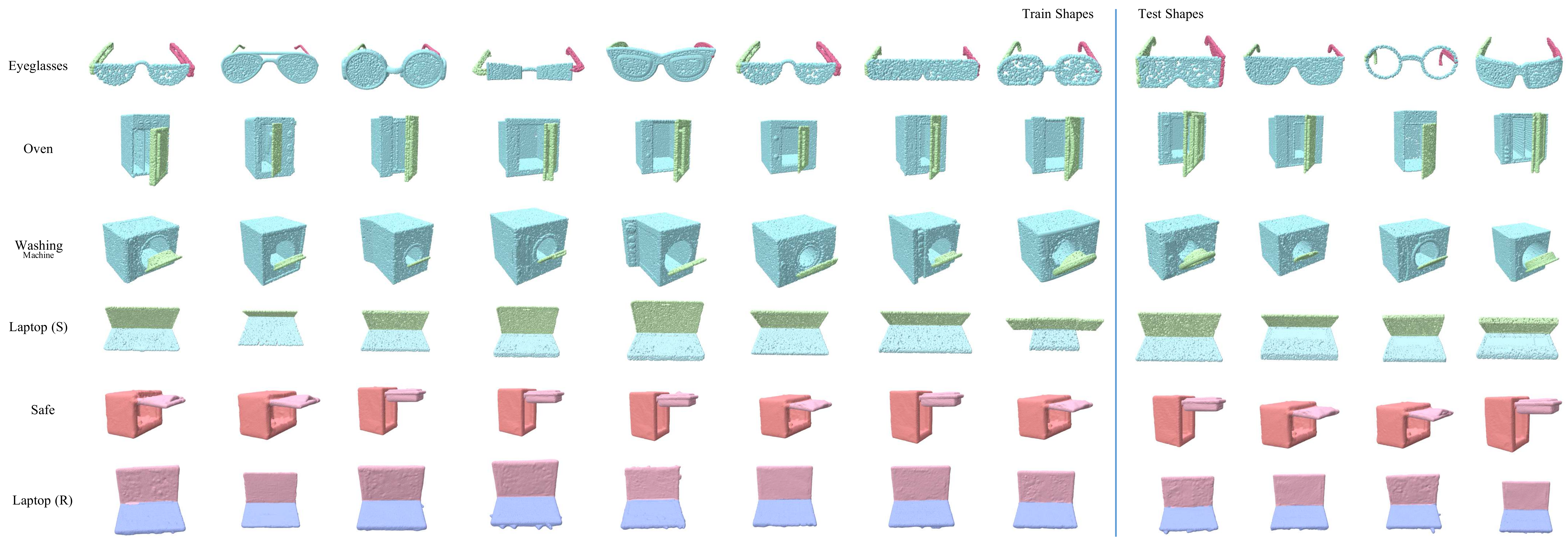}
    \vspace{-5pt}
  \caption{ \footnotesize
     Samples of training and test shapes.
  }
  \label{fig_data_train_test_split}
  % \vspace{-5pt}
  \vspace{-8pt}
\end{figure*}

\vpara{Data samples visualization.}
In Figure~\ref{fig_data_train_test_split}, we provide samples of training and test shapes for some categories for an intuitive understanding w.r.t. intra-category shape variations. 
Such variations mainly come from part geomtry (\emph{e.g.} Eyeglasses frames, Oven bodies, Laptop) and part size (\emph{e.g.} Washing Machine, Laptop).
% For instance, shape variation in the Eyeglasses category mainly lies in the geometry of eyeglass frames. 
% The variation of the Oven category lies in the shape of oven bodies. 
% Shape variations of the Washing Machine category come from the size and the geometric appearance of the body and the lid. 
% Shape variations of the Laptop category come from the size and the geometric appearance of the base and screen of laptop shapes.

\begin{table}[t]
  \centering
  \caption{\footnotesize 
  Per-category data splitting. 
%   100 samples in different articulation states are generated for each shape. For each articulated posed sample, 10 randomly rotated samples or 10 partial point clouds with different viewpoints are generated from it. 
  } 
% \vspace{-6pt}
\resizebox{0.8\linewidth}{!}{%
  \begin{tabular}{@{\;}c@{\;}|c|c|c|c|c|c|c@{\;}}
  \midrule
      \hline
      \specialrule{0em}{1pt}{0pt} 
      ~ & Oven & \makecell[c]{Washing \\ Machine} & Eyeglasses & Laptop (S) & Safe & Laptop (R) & Drawer \\ 
      \cline{1-8} 
      \specialrule{0em}{1pt}{0pt}
      
      \#Total & 32 & 41 & 42 & 82 & 30 & 50 & 30
      \\ 
      % \cline{1-8} 
      % \specialrule{0em}{1pt}{0pt}

      \#Train & 28 & 36 & 37 & 73 & 26 & 44 & 24
      \\ 
      % \cline{1-8} 
      % \specialrule{0em}{1pt}{0pt}

      \#Test & 4 & 5 & 5 & 9 & 4 & 6 & 6
      \\ \cline{1-8} 
      \specialrule{0em}{1pt}{0pt}
    
  \end{tabular}
  }
  % \vspace{-10pt}
  \vspace{-10pt}
  \label{tb_exp_dataset_meta_info}
\end{table}

% \todo{and other implementations for ICP}

\subsection{Implementation Details}  \label{sec_appen_imple_details}
% In this section, we introduce some implementation details for experiments. 
% The equivariant network backbone used here is EPN~\cite{chen2021equivariant}, 

\vpara{Architecture.}
For point convolution, we use a kernel-rotated version kernel point convolution (KPConv~\cite{thomas2019kpconv}) proposed in EPN~\cite{chen2021equivariant}. 
The size of the (one) convolution kernel is determined by the number of anchor points and the feature dimension. 
In our implementation, we use 24 anchor points. Feature dimensions at different convolution blocks are set to 64, 128, and 512 respectively.

\vpara{Training Protocol.}
In the training stage, the learning rate is set to $0.0001$, which is decayed by 0.7 every 1000 iterations. 
The model is trained for 10000 steps with batch size 8 on all datasets. 
We use the self-supervised reconstruction loss to train the network, with the weight for joint regularization $\lambda$ set to $1.0$ empirically. 
We use Adam optimizer with $\beta = (0.9, 0.999), \epsilon = 10^{-8}$.

\vpara{Software and Hardware Configurations.}
All models are implemented by PyTorch version 1.9.1, torch\_cluster version 1.5.1, torch\_scatter version 2.0.7, pyrender version 0.1.45, trimesh version 3.2.0, and Python 3.8.8. 
All the experiments are conducted on a Ubuntu 20.04.3 server with 8 NVIDIA 
% 24576MiB 
GPUs, 504G RAM, CUDA version 11.4. 

\subsection{Implementation details for baselines} \label{sec_appen_baselines}
% In this section, we explain some implementation details w.r.t. compared baselines. 

\vpara{NPCS~\cite{li2020category}.} 
The NPCS's original version~\cite{li2020category} 
% original version of NPCS proposed in~\cite{li2020category} 
trains a network for category-level articulated object pose estimation in a supervised manner.
It utilizes a PointNet++~\cite{qi2017pointnetpp} to regress three kinds of information and a set of  pre-defined normalized part coordinate spaces. 
Then in the evaluation process, the RANSAC algorithm is leveraged to calculate the rigid transformation of each part from its predicted normalized part coordinates to the shape in the camera space.  
To apply NPCS in our experiments, we make the following two modifications: 
1) We change the backbone used in NPCS from PointNet++ to EPN. 
We further add supervision on its rotation mode selection process for the major rotation matrix prediction as does in~\cite{chen2021equivariant}. 
2) We add a joint axis orientation prediction branch and a pivot point regression branch for joint parameters estimation. 
Such two prediction branches act on the global shape feature corresponding to the selected rotation mode and predict a residual rotation and a translation for estimation. 
By applying the major rotation matrix of the selected mode, we could then arrive at the joint axis orientations and pivot points in the camera space. 
% with the mode selection process supervisedly trained~\cite{chen2021equivariant}. 
% with arbitrary global pose variations
% where shapes may undergo an arbitrary global rigid transformation and also use NPCS to predict joint parameters, we make the following two improvements on the original version of NPCS:
% 1) We change the backbone used in NPCS from PointNet++ to EPN with the mode selection process supervisedly trained~\cite{chen2021equivariant}. 
% Specifically, we regress the feature in each mode to a score that is learned under the supervision of the ground-truth nearest mode. 
% 2) We add a joint axis direction and a pivot point regression branch to learn to predict the joint axis direction and pivot points from the feature of the selected mode. 
% After that, the rotation matrix of the selected mode is applied on the predicted result to get the joint axis direction and predicted pivot points of the observed shape. 

\vpara{Oracle ICP.} 
To apply ICP on the articulated object pose estimation, we introduce Oracle ICP. 
Oracle ICP registers each ground-truth part from the template shape to the observed shape iteratively. 
% For 
% This method denotes the strategy using template shapes and observed shapes with ground-truth segmentation for part-by-part registration. 
We randomly select 5 segmented shapes from the train set to register on each test shape.
For complete point clouds, we first centralize the part shape from both of the template shape and the observed shape, we then iteratively register the template part shape to the observed part shape under 60 initial hypotheses. 
For partial point clouds, we iteratively register the template part shape to the observed part shape under 60 initial hypotheses together with 10 initial translation hypotheses. 
The one that achieves the smallest inlier RMSE value is selected as the registration result.
After each registration, we assign per-point segmentation label as the label of the nearest part. 
Therefore, we also treat Oracle ICP as one of our segmentation baseline.

\vpara{BSP-Net~\cite{chen2020bsp}.}
BSP-Net reconstructs an input shape using implicit fields as the representation by learning to partition the shape using three levels of representations, from planes to convexes, and further to the concave. 
Indices of reconstructed covexes are consistent across different shapes in the same category. 
Thus, we can map from each convex index to a ground-truth segmentation label. 
% segmented convex indices to segmentation labels. 
The relationship can then help us get segmentations for test shapes. 
The mapping can then help us segment each test shape by assigning each convex to its corresponding part segment. 
The intra-category convex partition consistency further provide cross-instance aligned part segmentations. 
% However, due to the change of part articulation states and the global pose variation of our input shape, the mapping may not be consistent across different shapes. 
However, due to the global pose variations of our data, such convex index consistency may not be observed when directly applying BSP-Net on our data. 
Thus, we propose to improve the evaluation process of BSP-Net to mitigate this problem. 
% to tackle this problem. 
Specifically, for each test shape, we find a shape from the train set that is the most similar to the current test shape. 
Then, we directly use its convex-segmentation mapping relationship to get segments for the test shape.
The segments are then used to calculate the segmentation IoU for the test shape.

\vpara{NSD~\cite{kawana2020neural}.}
Neural Star Domain~\cite{kawana2020neural} decomposes shapes into parameterized primitives. 
To test its part segmentation performance on our data with arbitrary global pose variations, we adopt the evaluation strategy similar to that used for BSP-Net.

% For depth rendering, we first put camera at a fixed distance facing the object, the object is allowed to have small offsets and random rotations. For each category, we generate 30K training images with 6K testing images.

\subsection{Experiments on Partial Point Clouds} \label{sec_appen_exp_partial}
In this section, we present the experimental results on rendered partial point clouds of our method and baseline methods. 

\vpara{Articulated Object Pose Estimation.}
In Table~\ref{tb_exp_pose_cmp_partial}, we present the part pose estimation and joint parameter prediction results of our method and baseline methods.
Compared to the pose estimation results of different models achieved on complete point clouds, our model can sometimes outperform the \textbf{supervised} NPCS baseline (using EPN as the backbone), such as better part pose estimations and joint parameter prediction results on the Laptop (S) dataset. 
% rotation, translation, and joint prediction parameters' error of Laptop (S) datatset. 

\begin{figure*}[ht]
  \centering
    \includegraphics[width=1.0\textwidth]{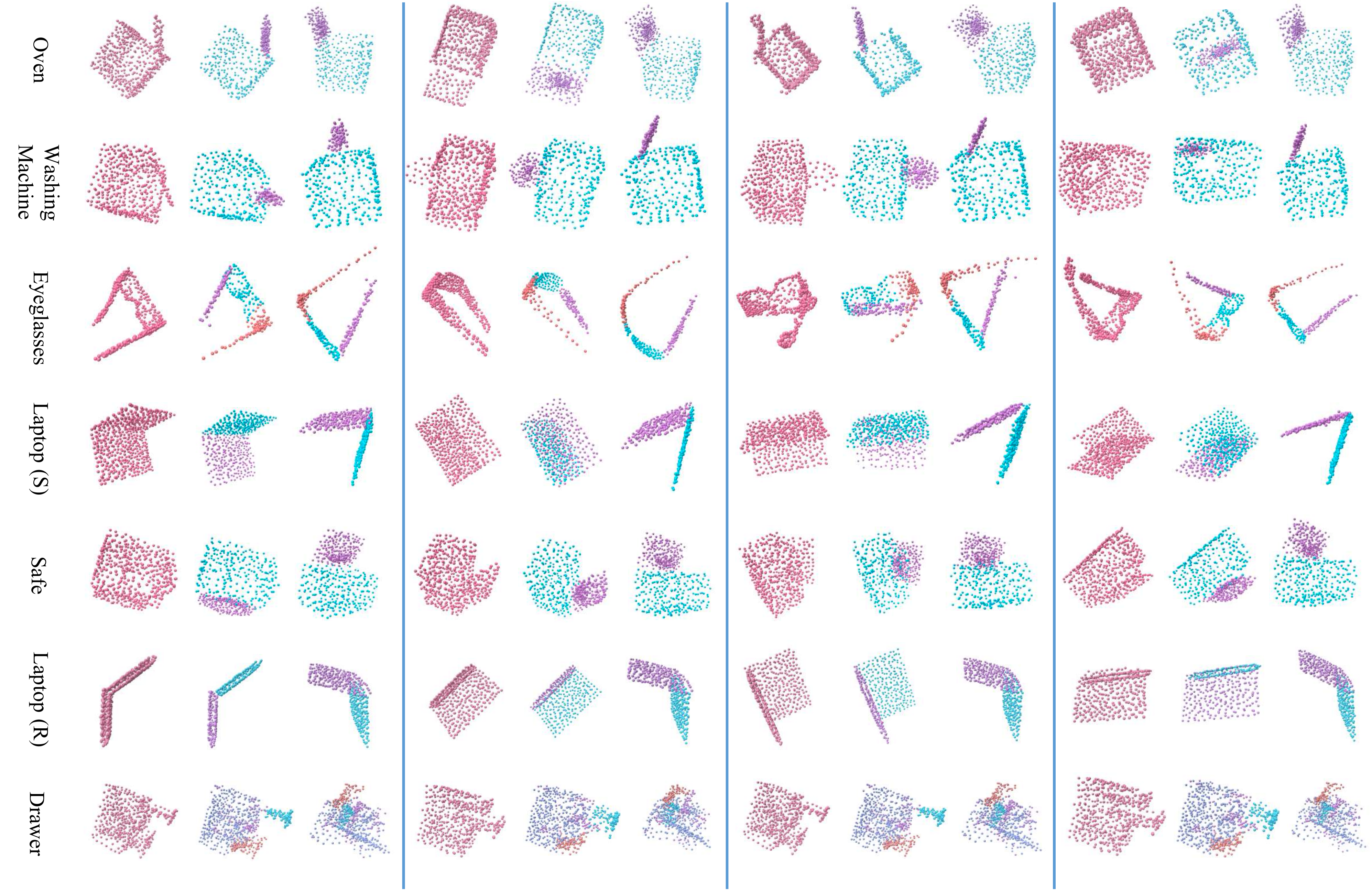}
    \vspace{-5pt}
  \caption{ \footnotesize
      Visualization for experimental results on complete point clouds. 
      Shapes drawn for every three shapes from the left side to the right side are the input point cloud, reconstructions, and the predicted canonical object shape. 
      \textbf{We put drawers in an aligned space just for better visualization.} 
      Their global pose may vary when feeding into the network.  Please zoom in for details. 
    %   There will be an arbitrary SE(3) transformation added on input shapes for drawers as well. Please zoom in for details. 
    %   We can see that the lid 
  }
  \label{fig_complete_vis}
  % \vspace{-5pt}
  \vspace{-8pt}
\end{figure*}

\begin{figure*}[ht]
  \centering
    \includegraphics[width=1.0\textwidth]{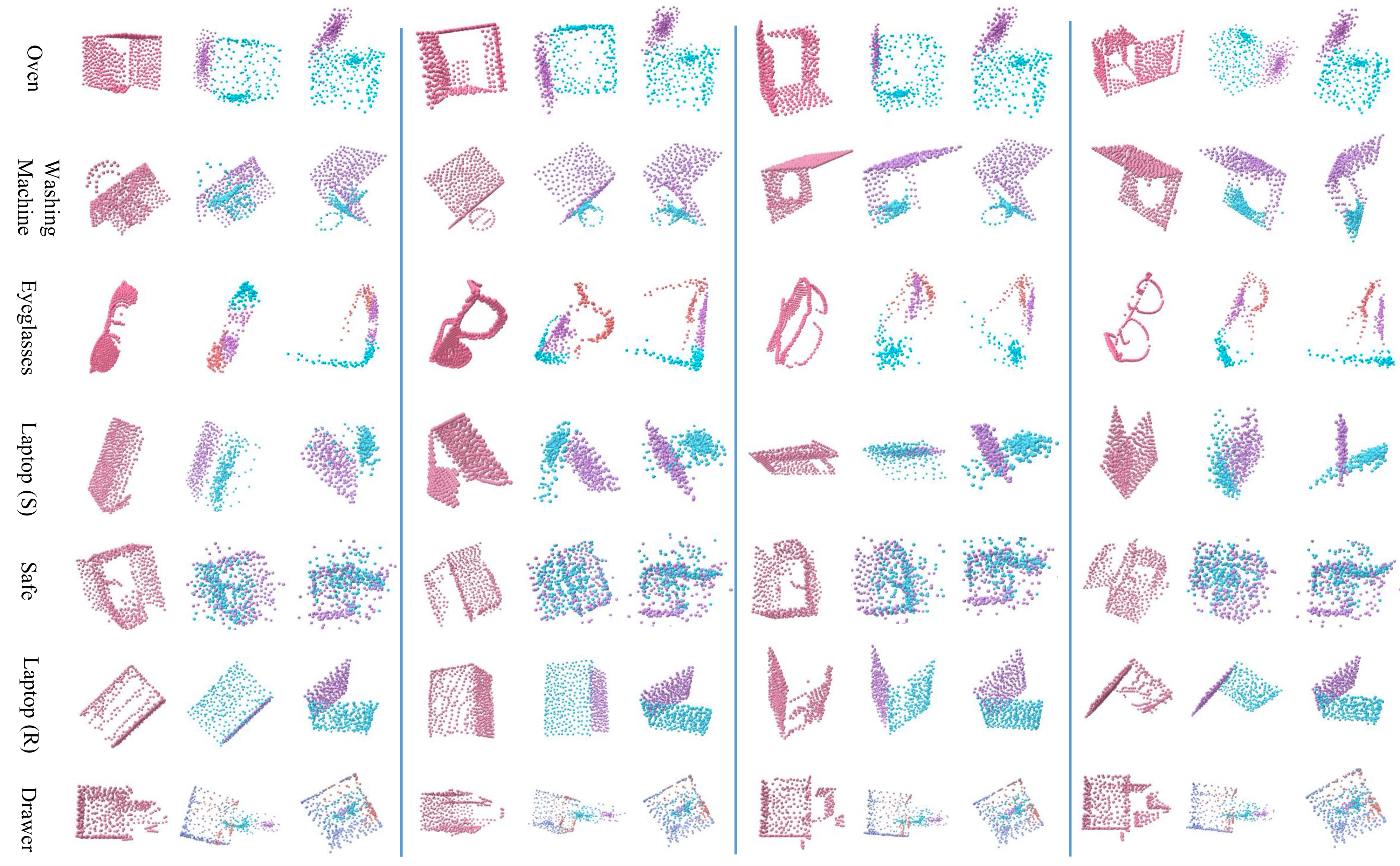}
    \vspace{-5pt}
  \caption{ \footnotesize
      Visualization for experimental results on \textbf{\textcolor{red}{partial point clouds}}. 
      Shapes drawn for every three shapes from the left side to the right side are the input point cloud, reconstructions, and the predicted canonical object shape. Please zoom in for details. 
  }
  \label{fig_partial_vis}
  % \vspace{-5pt}
  \vspace{-8pt}
\end{figure*}

In Figure~\ref{fig_partial_vis}, we draw some samples for a qualitative evaluation. 
Moreover, we also provide the visualization of all categories for complete point clouds in Figure~\ref{fig_complete_vis}.
In the figure, the point cloud distance function used for Safe is unidirectional Chamfer Distance, while that used for others is still bidirectional Chamfer Distance. 
Using unidirectional Chamfer Distance can relieve the problem of joint regularization on partial point clouds to some extent. 
It is because that in this way the point cloud completion could be naturally enforced. 
For instance, reconstruction results for the Safe category are drawn in Figure~\ref{fig_partial_vis}. 
However, points that are not mapped to any point in the input shape will also affect the point-based joint regularization. 
For simple shapes, using bidirectional Chamfer Distance could also sometimes make the decoder decode complete part shapes, e.g. reconstructions for Laptop (R). 
As for the reconstructed reference shape in the canonical object space, better joint predictions would lead to better global shape alignment. 
% good global shape alignment and articulation state normalization could be observed if the joint prediction result is good. 
For instance, we can observe that the angle between two parts of Laptop (R) and Laptop (S) is relatively consistent across shapes with different articulation states. 
Joint regularization enforcing the connectivity between two adjacent parts both before and after the articulated transformation in the canonical object space.
It could then help make the joint behave like a real joint, based on which we the ``lazy'' network tends to decode part shapes with consistent orientations. 
However, there is a degenerated solution where the decoded rotation angle is put near to zero. 
In that case, the joint regularization term could be satisfied by decoding ``twisted'' part shapes. 
Since the decoded angle is near zero, the connectivity between two parts will not be broken when rotating along the decoded joint. 
Sometimes, decoding angles near to zeros is a local minimum that the optimization process gets stuck in. 
At that time, the regularization loss term is a large one but the decoded joint parameters are not optimized in the correct direction by the network. 
The reconstructed shapes in the canonical object space do not have consistent angles, e.g. Washing Machine and Safe drawn in Figure~\ref{fig_partial_vis}.

\vpara{Part Segmentation.}
In Table~\ref{tb_exp_seg_cmp_partial}, we evaluate the segmentation performance of our method. 
BSP-Net is not compared here since it requires mesh data for pre-processing, which is not compatible with the rendered partial point clouds. 
% due to the data preparation process that needs mesh data as input, which is not compatible with the rendered partial point clouds.
Oracle ICP uses real segmentation labels to register each part from the example shape to the observed shape. 
% Thus, it is not surprising that it can achieve good performance on tested categories. 
% However, due to the occluded shape, and the ambiguity caused by occluded part shapes and part symmetry, the part pose estimation performance of ICP is not that satisfactory.
Despite this, it can still not achieve satisfactory estimation results due to shape occlusions and part-symmetry-related pose ambiguity issues. 

\vpara{Shape Reconstruction.}
In Table~\ref{tb_exp_completion_cmp_partial}, we evaluate the shape reconstruction performance of our method. 
The part-by-part reconstruction strategy used by our method can outperform the EPN-based whole shape reconstruction strategy in most of those categories except for Washing Machine. 
One possible reason is the poor segmentation performance of our model on shapes in the Washing Machine category.

\begin{table}[t]
  \centering
  \caption{\footnotesize 
  Comparison between the part pose estimation performance of different methods on all test categories (\textbf{\textcolor{red}{partial point clouds}}). 
  ``R'' denotes rotation errors with the value format  
    % , whose values are 
    % % presented for rotation errors 
    % in the format of 
    ``Mean $R_{err}$/Median $R_{err}$''. ``T'' denotes translation errors with the value format 
    % , whose values 
    % % presented 
    % are in the format of 
    ``Mean $T_{err}$/Median $T_{err}$''. ``J'' denotes joint parameters estimation results with the value format ``Mean $\theta_{err}$/Mean $d_{err}$''.
%   ``R'' denotes rotation errors with values are 
%   in the format of ``Mean $R_{err}$/Median $R_{err}$''. 
%   ``T'' denotes translation errors, whose values 
%   are in the format of ``Mean $T_{err}$/Median $T_{err}$''. ``J'' denotes joint parameters-related estimation errors,whose values 
%   are in the format of ``Mean $\theta_{err}$/Mean $d_{err}$''. 
\textbf{ICP could not predict joint parameters.} Therefore, only the results of supervised NPCS and our method on joint prediction are presented.
  For all metrics, the smaller, the better. \textbf{Bold} numbers for best values, while \emph{\textcolor{blue}{blue}} values represent second best ones. 
  } 
%   \vspace{-8pt}
\resizebox{\linewidth}{!}{%
  \begin{tabular}{@{\;}c@{\;}|c|c|c|c|c|c|c|c|c@{\;}}
  \midrule
      \hline
      \specialrule{0em}{1pt}{0pt} 
      ~ & Method & Oven & \makecell[c]{Washing \\ Machine} & Eyeglasses & Laptop (S) & Safe & Laptop (R) & Drawer & Avg. \\ 
      \cline{1-10} 
      \specialrule{0em}{1pt}{0pt}
      
      % mean error / Median error
      \multirow{6}{*}{R} & \makecell[c]{NPCS-EPN \\ (supervised)}   & \makecell[c]{\textbf{2.51}/\textbf{2.27},  \\ \textbf{{2.93}}/\textbf{{2.64}}} & \makecell[c]{\emph{\textcolor{blue}{4.71}}/\emph{\textcolor{blue}{3.84}}, \\ \textbf{8.56}/\textbf{7.46}} & \makecell[c]{\emph{\textcolor{blue}{7.26}}/\emph{\textcolor{blue}{6.08}}, \\ \emph{\textcolor{blue}{23.39}}/\emph{\textcolor{blue}{17.33}}, \\ \emph{\textcolor{blue}{20.86}}/\emph{\textcolor{blue}{18.76}}} & \makecell[c]{\emph{\textcolor{blue}{21.40}}/\emph{\textcolor{blue}{23.56}}, \\ \emph{\textcolor{blue}{29.90}}/\emph{\textcolor{blue}{32.71}}} & \makecell[c]{\textbf{6.64}/\textbf{5.76}, \\ \textbf{5.43}/\textbf{5.19}} & \makecell[c]{\emph{\textcolor{blue}{9.39}}/\emph{\textcolor{blue}{8.75}}, \\ \textbf{{6.75}}/\emph{\textcolor{blue}{6.14}}} & \makecell[c]{\textbf{21.74}/\textbf{10.80}, \\ \textbf{22.92}/\textbf{10.18}, \\ \textbf{25.10}/\textbf{14.16}, \\ \textbf{7.34}/\textbf{6.83}} & \textbf{13.34}/\textbf{10.73}
      \\ \cline{2-10} 
      \specialrule{0em}{1pt}{0pt}

      ~ & Oracle ICP & \makecell[c]{{21.53}/10.80,  \\ {20.68}/20.50} & \makecell[c]{32.42/17.82, \\ 19.39/16.99} & \makecell[c]{73.24/78.73, \\ 68.74/74.09, \\ 69.23/74.53} & \makecell[c]{67.48/73.01, \\ 63.22/68.23} & \makecell[c]{38.72/34.44,\\ 52.28/42.16} & \makecell[c]{30.78/28.674, \\ 42.06/39.25} & \makecell[c]{82.93/82.64,\\61.31/59.51,\\54.39/52.82,\\26.88/29.66} & 48.55/47.29
      \\ 
      \cline{2-10} 
      \specialrule{0em}{1pt}{0pt}

      % mean error / Median error
      ~ & Ours & \makecell[c]{\emph{\textcolor{blue}{11.77}}/\emph{\textcolor{blue}{7.87}},  \\ \emph{\textcolor{blue}{10.83}}/\emph{\textcolor{blue}{9.15}}} & \makecell[c]{\textbf{{1.61}}/\textbf{{1.52}}, \\ \emph{\textcolor{blue}{12.81}}/\emph{\textcolor{blue}{12.51}}} & \makecell[c]{\textbf{{4.69}}/\textbf{{3.77}}, \\ \textbf{{9.56}}/\textbf{{5.36}}, \\ \textbf{{7.53}}/\textbf{{6.12}}} & \makecell[c]{\textbf{{10.18}}/\textbf{5.30}, \\ \textbf{11.10}/\textbf{5.22}} & \makecell[c]{\emph{\textcolor{blue}{15.38}}/\emph{\textcolor{blue}{14.46}}, \\ \emph{\textcolor{blue}{21.91}}/\emph{\textcolor{blue}{19.04}}} & \makecell[c]{\textbf{8.50}/\textbf{6.85}, \\ \emph{\textcolor{blue}{6.92}}/\textbf{5.66}} & \makecell[c]{\emph{\textcolor{blue}{2.60}}/\emph{\textcolor{blue}{1.79}}, \\ \emph{\textcolor{blue}{2.60}}/\emph{\textcolor{blue}{1.79}}, \\ \emph{\textcolor{blue}{2.06}}/\emph{\textcolor{blue}{1.79}}, \\ \emph{\textcolor{blue}{2.06}}/\emph{\textcolor{blue}{1.79}}} & \emph{\textcolor{blue}{8.36}}/\emph{\textcolor{blue}{6.47}}
      \\ \cline{1-10} 
      \specialrule{0em}{1pt}{0pt}
      
      \multirow{6}{*}{T} & \makecell[c]{NPCS-EPN \\ (supervised)}  & \makecell[c]{\textbf{0.028}/\textbf{0.030}, \\ \textbf{0.028}/\textbf{0.023}} & \makecell[c]{\textbf{0.034}/\textbf{0.030}, \\ \textbf{0.033}/\textbf{{0.028}}}  & \makecell[c]{\textbf{0.085}/\textbf{0.075}, \\ \textbf{0.056}/\textbf{0.052}, \\ \textbf{0.057}/{\textbf{0.049}}} & \makecell[c]{\emph{\textcolor{blue}{0.263}}/\emph{\textcolor{blue}{0.253}}, \\ {0.286}/{0.236}} & \makecell[c]{\textbf{0.022}/\textbf{0.021}, \\ \textbf{0.034}/\textbf{0.034}} & \makecell[c]{\textbf{0.048}/\textbf{0.043}, \\ \textbf{0.047}/\textbf{0.044}} & \makecell[c]{{\textbf{0.441}}/\emph{\textcolor{blue}{0.365}}, \\ \textbf{0.367}/\emph{\textcolor{blue}{0.343}}, \\ \textbf{0.549}/\textbf{0.299}, \\ \textbf{0.081}/\textbf{0.065}} & \textbf{0.145}/\emph{\textcolor{blue}{0.117}}
      \\ \cline{2-10} 
      \specialrule{0em}{1pt}{0pt}

      ~ & Oracle ICP & \makecell[c]{{0.324}/{{0.321}},  \\ \emph{\textcolor{blue}{0.169}}/\emph{\textcolor{blue}{0.171}}} & \makecell[c]{0.322/{{0.311}}, \\ \emph{\textcolor{blue}{0.136}}/\emph{\textcolor{blue}{0.144}}} & \makecell[c]{\emph{\textcolor{blue}{0.092}}/\emph{\textcolor{blue}{0.097}}, \\ 0.188/0.197, \\ 0.185/0.193} & \makecell[c]{0.265/0.278, \\ \emph{\textcolor{blue}{0.267}}/\emph{\textcolor{blue}{0.277}}} & \makecell[c]{0.281/{{0.280}}, \\ 0.246/{{0.248}}} & \makecell[c]{0.280/0.289,\\ 0.305/0.306} & \makecell[c]{\emph{\textcolor{blue}{0.193}}/\emph{\textcolor{blue}{0.197}},\\ \emph{\textcolor{blue}{0.161}}/\emph{\textcolor{blue}{0.170}}, \\ \emph{\textcolor{blue}{0.159}}/\textbf{0.164},\\\emph{\textcolor{blue}{0.129}}/{0.132}} & 0.218/0.222
      \\ 
      \cline{2-10} 
      \specialrule{0em}{1pt}{0pt}

      ~  & Ours & \makecell[c]{\emph{\textcolor{blue}{0.071}}/\emph{\textcolor{blue}{0.065}}, \\ {{0.204}}/\emph{\textcolor{blue}{0.120}}} & \makecell[c]{\emph{\textcolor{blue}{0.179}}/\emph{\textcolor{blue}{0.164}}, \\ 0.253 /0.254}  & \makecell[c]{{{ 0.219}}/{{0.226}}, \\ \emph{\textcolor{blue}{0.169}}/\emph{\textcolor{blue}{0.166}}, \\ \emph{\textcolor{blue}{0.177}}/\emph{\textcolor{blue}{0.171}}} & \makecell[c]{{\textbf{0.044}}/\textbf{{0.034}}, \\ \textbf{{0.031}}/\textbf{0.025}} & \makecell[c]{\emph{\textcolor{blue}{0.030}}/\emph{\textcolor{blue}{0.030}}, \\ \emph{\textcolor{blue}{0.100}}/\emph{\textcolor{blue}{0.104}}} & \makecell[c]{\emph{\textcolor{blue}{0.088}}/\emph{\textcolor{blue}{0.082}}, \\ \emph{\textcolor{blue}{0.070}}/\emph{\textcolor{blue}{0.067}}} & \makecell[c]{{0.046}/{0.046}, \\ {0.047}/{0.050}, \\ {0.122}/{0.131}, \\ 0.172/0.142} & \emph{\textcolor{blue}{0.119}}/\textbf{0.110}
      \\ \cline{1-10} 
      \specialrule{0em}{1pt}{0pt}
      
      \multirow{2}{*}{J} & \makecell[c]{NPCS-EPN \\ (supervised)}  & {28.62}/\textbf{0.092} & \textbf{8.05}/\textbf{0.194}  & \makecell[c]{\textbf{20.11}/0.221,\\ \textbf{20.11}/\textbf{0.239}}   & {10.91}/0.155 & \textbf{11.23}/\textbf{0.084 } & \textbf{12.25}/\textbf{0.134} & {11.21}/- & \textbf{15.31}/\textbf{0.160}
      \\ \cline{2-10} 
      \specialrule{0em}{1pt}{0pt}

      ~ & Ours & \textbf{5.24}/{0.105 } & 22.30/0.212  &\makecell[c]{26.96/\textbf{0.087},\\ 26.96/0.260} & \textbf{10.83}/\textbf{0.142} & 55.16/0.170 & 18.02/0.170  & \textbf{7.43}/- & 21.61/0.164
      \\ \cline{1-10} 
      \specialrule{0em}{1pt}{0pt}

  \end{tabular}
  }
  % \vspace{-14pt}
  \label{tb_exp_pose_cmp_partial}
\end{table} 

\begin{table}[t]
  \centering
  \caption{\footnotesize 
  Comparison between the part segmentation performance of different methods (\textbf{\textcolor{red}{partial point clouds}}). The metric used for this task is Segmentation MIoU, calculated on 4096 points for each shape. Values presented in the table are scaled by 100. Larger values indicate better performance. 
  } 
% \vspace{-8pt}
\resizebox{0.8\linewidth}{!}{%
  \begin{tabular}{@{\;}c@{\;}|c|c|c|c|c|c|c@{\;}}
  \midrule
      \hline
      \specialrule{0em}{1pt}{0pt} 
      ~ & Oven & \makecell[c]{Washing\\ Machine} & Eyeglasses & Laptop (S) & Safe & Laptop (R) & Drawer
      \\ 
      \cline{1-8} 
      \specialrule{0em}{1pt}{0pt}

      % mean error / Median error
      Oracle ICP & 75.83  & \textbf{73.07} & \textbf{68.92} & 54.01 & \textbf{66.90}  & 59.96 & \textbf{58.38}
      \\ 
      % \cline{1-8} 
      % \specialrule{0em}{1pt}{0pt}

      % mean error / Median error
      Ours & \textbf{87.07}  & 51.73 & 56.80 & \textbf{84.94} & 44.64 & \textbf{86.04} & {45.45}
      \\ \cline{1-8} 
      \specialrule{0em}{1pt}{0pt}
      
  \end{tabular}
  }
  %  \vspace{-10pt}
    \vspace{-12pt}
  \label{tb_exp_seg_cmp_partial}
\end{table} 

\begin{table}[t]
  \centering
  \caption{\footnotesize 
  Comparison between the shape reconstruction performance of different methods (\textbf{\textcolor{red}{partial point clouds}}). The metric used in this task is unidirectional Chamfer L1 from the original input shape to the reconstructed shape. The smaller, the better. 
  } 
% \vspace{-8pt}
\resizebox{0.8\linewidth}{!}{%
  \begin{tabular}{@{\;}c@{\;}|c|c|c|c|c|c|c@{\;}}
  \midrule
      \hline
      \specialrule{0em}{1pt}{0pt} 
      Method & Oven & \makecell[c]{Washing \\ Machine} & Eyeglasses & Laptop (S) & Safe & Laptop (R) & Drawer \\ 
      \cline{1-8} 
      \specialrule{0em}{1pt}{0pt}
      
      % mean error / Median error
      EPN~\cite{li2021leveraging} & {0.040} & \textbf{0.043} & 0.044 & 0.032  & 0.020 & 0.026 &  0.079
      \\
      % \cline{1-8}
      % \specialrule{0em}{1pt}{0pt}
      
      Ours & \textbf{0.035} & 0.062 & \textbf{0.041} & \textbf{0.025} & \textbf{0.019} & \textbf{0.024} & \textbf{0.061}
      \\ \cline{1-8}
      \specialrule{0em}{1pt}{0pt}
    
  \end{tabular}
  }
  % \vspace{-10pt}
  \vspace{-10pt}
  \label{tb_exp_completion_cmp_partial}
\end{table} 

\subsection{Additional Comparisons and Applications} \label{sec_appen_more_exp_additional}

\vpara{Comparison with Other Baselines.}
%%% EPN and NPCS
We compare our method with other two baselines that are not discussed in the main body in Table~\ref{tb_exp_pose_cmp_other_baseline}. 

Firstly, we use KPConv~\cite{thomas2019kpconv} as NPCS's feature backbone (denoted as ``NPCS-KPConv'') and test its performance on our data with arbitrary global pose variation. 
We can see that NPCS of this version performs terribly compared to our unsupervised method. 
NPCS estimates part poses by estimating the transformation from estimated NPCS coordinates and the observed shape. 
It therefore requires invariant NPCS predictions to estimate category-level part poses. 
However, such prediction consistency may not be easily achieved for input shapes with various global pose variations. 
% requiring the invariance that the shape reconstruction network should achieve when the input shape undergoes an arbitrary SE(3) transformation. 
% The poor performance achieved by KPConv may reveal the effectiveness of using equivariant networks for pose and shape factorization. 
% KPConv's poor performance may further demonstrate the necessary of using equivariant netwo

The second one is Oracle EPN, where we assume ground-truth part segmentation labels and use EPN to estimate the pose for each individual part. 
Despite in such oracle setting, EPN cannot infer joint parameters since it estimates per-part poses individually. 
% they estimate the pose of each part individually. 
Besides, the part symmetry problem will also hinder such strategy from getting good performance to some extent, which will be discussed in the next section~\ref{sec_appen_symmetric_parts}. 
% \todo{xxx}

\begin{figure*}[ht]
  \centering
    \includegraphics[width=1.0\textwidth]{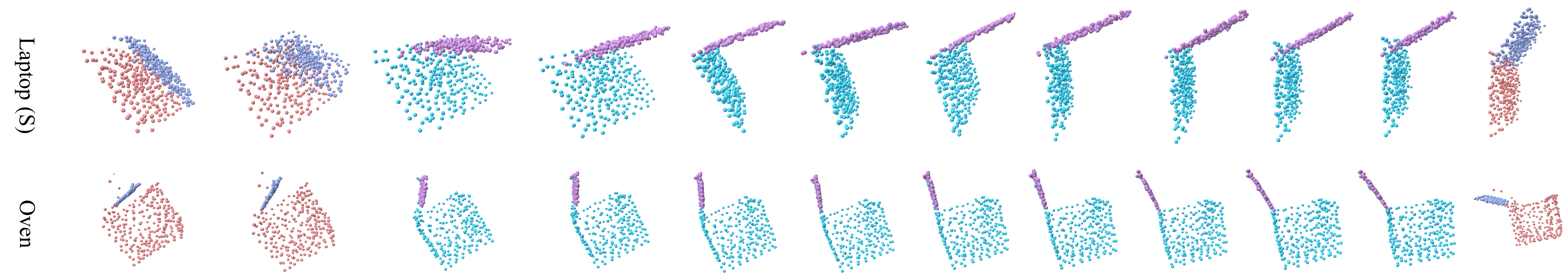}
    \vspace{-5pt}
  \caption{ \footnotesize
    Reconstruction for shapes in different articulation states and manipulations to change their states.
    % Predicted 
    % Reconstruction for shapes in different articulation states and manipulation to change their states. 
    Shapes (in blue and orange) drawn on the two sides are manipulated shapes from their nearest reconstructions. 
    Others are reconstructions (in purple and green). Please zoom in for details. 
  }
  \label{fig_mani_vis}
  % \vspace{-5pt}
  \vspace{-8pt}
\end{figure*}

\vpara{Shape Reconstruction and Manipulation.}
The predicted joints can enable us to manipulate the reconstruction by changing the value of predicted rotation angles. 
We then arrive at shapes in new sarticulation states different from input shapes. 
% The resulting shapes then undergo a change in the articulation state from the input shape. 
% shape manipulation based on the segmentation/reconstruction of one single shape to change its articulation states. 
In Figure~\ref{fig_mani_vis}, we draw some examples for Laptop (S) and Oven.

\begin{table}[t]
  \centering
  \caption{\footnotesize 
  Comparison between the part pose estimation performance of different methods. 
  Backbone used for NPCS is KPConv. 
  ``R'' denotes rotation errors with the value format  
    % , whose values are 
    % % presented for rotation errors 
    % in the format of 
    ``Mean $R_{err}$/Median $R_{err}$''. ``T'' denotes translation errors with the value format 
    % , whose values 
    % % presented 
    % are in the format of 
    ``Mean $T_{err}$/Median $T_{err}$''. ``J'' denotes joint parameters estimation results with the value format ``Mean $\theta_{err}$/Mean $d_{err}$''.
  For all metrics, the smaller, the better. \textbf{{Bold}} numbers for best values. 
  } 
%   \vspace{-8pt}
\resizebox{\linewidth}{!}{%
  \begin{tabular}{@{\;}c@{\;}|c|c|c|c|c|c|c|c|c@{\;}}
  \midrule
      \hline
      \specialrule{0em}{1pt}{0pt} 
      ~ & Method & Oven & \makecell[c]{Washing \\ Machine} & Eyeglasses & Laptop (S) & Safe & Laptop (R) & Drawer & Avg. \\ 
      \cline{1-10} 
      \specialrule{0em}{1pt}{0pt}
      
      % mean error / Median error
      \multirow{6}{*}{R} & \makecell[c]{NPCS-KPConv \\ (supervised)}  & \makecell[c]{{44.16}/{43.09},  \\ {{60.58}}/{{63.35}}} & \makecell[c]{56.20/56.22,\\50.16/51.38}  & \makecell[c]{51.99/53.97,\\ 42.48/38.08, \\ 42.29/38.11} & \makecell[c]{55.67/66.44,\\55.63/61.33} & \makecell[c]{11.68/11.10,\\ 43.48/42.22} & \makecell[c]{49.98/68.43,\\ 73.40/83.55} & \makecell[c]{62.73/69.42,\\ 56.16/60.34, \\ 57.23/63.90, \\ 48.76/46.82} & 50.74/53.99
      \\ \cline{2-10} 
      \specialrule{0em}{1pt}{0pt}
      
      % mean error / Median error
      ~ & Oracle EPN & \makecell[c]{\textbf{{7.07}}/\textbf{{6.88}},  \\ {16.33}/{9.17}} & \makecell[c]{{{7.97}}/{{7.60}}, \\ {{33.56}}/{{20.49}}} & \makecell[c]{{{54.01}}/{{13.09}}, \\ {{86.12}}/{{65.07}}, \\ {{116.56}}/{{119.23}}} &  \makecell[c]{{{18.33}}/{9.73}, \\ {18.98}/{12.75}} & \makecell[c]{{{45.85}}/{{48.59}}, \\ {{38.03}}/{{27.67}}} & \makecell[c]{{20.46}/{14.03}, \\ {21.08}/{19.30}} & \makecell[c]{{{47.88}}/{{47.03}}, \\ {{30.84}}/{{25.23}}, \\ {{35.79}}/{{37.17}}, \\ {{43.83}}/{{39.46}}}  & 37.81/30.73
      \\ \cline{2-10} 
      \specialrule{0em}{1pt}{0pt}

      % mean error / Median error
      ~ & Ours & \makecell[c]{{{7.74}}/{{7.35}},  \\ \textbf{{4.07}}/\textbf{{3.97}}} & \makecell[c]{\textbf{{7.49}}/\textbf{{7.37}}, \\ \textbf{{19.27}}/\textbf{{19.19}}} & \makecell[c]{\textbf{{8.16}}/\textbf{{8.21}}, \\ \textbf{{12.29}}/\textbf{{10.89}}, \\ \textbf{{12.53}}/\textbf{{9.88}}} & \makecell[c]{\textbf{{7.34}}/\textbf{5.16}, \\ \textbf{10.41}/\textbf{9.34}} & \makecell[c]{\textbf{{9.03}}/\textbf{{9.09}}, \\ \textbf{{13.83}}/\textbf{{13.59}}} & \makecell[c]{\textbf{5.71}/\textbf{3.61}, \\ \textbf{3.64}/\textbf{2.84}} & \makecell[c]{\textbf{{3.18}}/\textbf{{2.73}}, \\ \textbf{{3.18}}/\textbf{{2.73}}, \\ \textbf{{3.18}}/\textbf{{2.71}}, \\ \textbf{{3.18}}/\textbf{{2.71}}}  & \textbf{7.90}/\textbf{7.14}
      \\ \cline{1-10} 
      \specialrule{0em}{1pt}{0pt}
      
      \multirow{6}{*}{T} & \makecell[c]{NPCS-KPConv \\ (supervised)}  & \makecell[c]{{0.133}/{0.121}, \\ {0.104}/0.091} & \makecell[c]{0.146/0.142,\\ 0.066/0.065}  & \makecell[c]{{0.401}/{0.326}, \\ {0.418}/{0.257}, \\ {0.396}/{{0.263}}} & \makecell[c]{0.233/0.203,\\ 0.217/0.169} & \makecell[c]{\textbf{0.055}/\textbf{0.052}, \\ {0.098}/{0.091}} & \makecell[c]{0.179/0.226,\\ 0.161/0.174} & \makecell[c]{{{0.791}}/{{0.742}}, \\ {0.694}/{{0.640}}, \\ {1.005}/{0.942}, \\ {0.271}/{0.240}}  & 0.316/0.279
      \\ \cline{2-10} 
      \specialrule{0em}{1pt}{0pt}
      
      ~  & Oracle EPN & \makecell[c]{\textbf{{0.031}}/\textbf{{0.030}}, \\ \textbf{{0.058}}/{{0.052}}} & \makecell[c]{\textbf{{0.046}}/\textbf{{0.044}}, \\ {0.059}/{0.053}}  & \makecell[c]{{{0.197}}/{{0.129}}, \\ {{0.128}}/{{0.118}}, \\ {{0.334}}/{{0.292}}} & \makecell[c]{{{0.132}}/{{0.128}}, \\ {{0.117}}/{0.090}} & \makecell[c]{{{0.157}}/0.157, \\ {{0.158}}/{0.151}} & \makecell[c]{\textbf{{0.092}}/\textbf{{0.086}}, \\ \textbf{{0.094}}/\textbf{{0.082}}} & \makecell[c]{{0.204}/{0.187}, \\ {0.177}/{0.166}, \\ {0.161}/{0.146}, \\ { 0.290}/{0.282}}  & 0.143/0.129
      \\ \cline{2-10} 
      \specialrule{0em}{1pt}{0pt}

      ~  & Ours & \makecell[c]{{{0.054}}/{{0.052}}, \\ {{0.067}}/\textbf{{0.046}}} & \makecell[c]{\textbf{{0.082}}/\textbf{{0.083}}, \\ \textbf{{0.042}}/\textbf{{0.034}}}  & \makecell[c]{{\textbf{{0.054}}}/\textbf{{0.039}}, \\ \textbf{{0.086}}/\textbf{{0.088}}, \\ \textbf{{0.070}}/\textbf{{0.055}}} & \makecell[c]{{\textbf{0.040}}/\textbf{{0.037}}, \\ \textbf{{0.046}}/\textbf{0.042}} & \makecell[c]{{{0.066}}/0.069, \\ \textbf{{0.037}}/\textbf{0.035}} & \makecell[c]{\textbf{{0.021}}/\textbf{{0.019}}, \\ \textbf{{0.027}}/\textbf{{0.026}}} & \makecell[c]{\textbf{0.096}/\textbf{0.096}, \\ \textbf{0.097}/\textbf{0.092}, \\ \textbf{0.108}/\textbf{0.105}, \\ \textbf{0.109}/\textbf{0.100}}  & \textbf{0.065}/\textbf{0.060}
      \\ \cline{1-10} 
      \specialrule{0em}{1pt}{0pt}
      
      \multirow{2}{*}{J} & \makecell[c]{NPCS-KPConv \\ (supervised)}  & {55.62}/{0.194} & 55.01/0.149  & \makecell[c]{{60.58}/0.329,\\ {60.59}/{0.379}} & 41.40/0.259 & 54.07/0.055 & 57.04/0.070 & 52.48/-  & 54.60/0.205
      \\ \cline{2-10} 
      \specialrule{0em}{1pt}{0pt}

      ~ & Ours & \textbf{20.30}/\textbf{0.089} & \textbf{28.40}/\textbf{0.118}  & \makecell[c]{\textbf{17.75}/\textbf{0.045},\\ \textbf{17.75}/\textbf{0.129}}  & \textbf{30.31}/\textbf{0.122} & \textbf{4.36}/\textbf{0.031} & \textbf{17.17}/\textbf{0.169} & \textbf{38.86}/-  & \textbf{21.86}/\textbf{0.100}
      \\ \cline{1-10} 
      \specialrule{0em}{1pt}{0pt}
    
  \end{tabular}
  }
  % \vspace{-14pt}
  \label{tb_exp_pose_cmp_other_baseline}
\end{table}

\subsection{Robustness to Input Data Noise} \label{sec_appen_robustness}

\begin{table}[t]
  \centering
  \caption{\footnotesize 
  Performance comparison of the proposed method on clean data and data corrupted by random normal noise. 
  }  
%   \vspace{-8pt}
\resizebox{1.0\linewidth}{!}{%
  \begin{tabular}{@{\;}c@{\;}|c|c|c|c|c|c|c|c@{\;}}
  \midrule
      \hline
      \specialrule{0em}{1pt}{0pt} 
      Category & Method & Seg. IoU & Mean $R_{err}(^\circ)$ & Median $R_{err}(^\circ)$ & Mean $T_{err}$ & Median $T_{err}$ & Joint Error & Chamfer L1 
      \\ 
      \cline{1-9} 
      \specialrule{0em}{1pt}{0pt}
      
      \multirow{2}{*}{Oven} & Without noise & \textbf{76.22} & \textbf{7.74}, \textbf{4.07} & \textbf{7.35}, \textbf{3.97} & \textbf{0.054}, {0.067} & \textbf{0.052}, \textbf{0.046}  & 20.30/\textbf{0.089}  & \textbf{0.025} 
      \\ \cline{2-9} 
      \specialrule{0em}{1pt}{0pt}
      
      ~ & With noise & {55.35} & {9.84}, {11.05} & {9.94}, {9.99} & {0.073}, \textbf{0.063} & {0.073}, {0.057} & \textbf{9.28}/0.310   & 0.049
      \\ \cline{1-9} 
      \specialrule{0em}{1pt}{0pt}
      
      % 82.97  & {7.34, 10.41} & {5.16, 9.34} & 0.040, 0.046 & {0.037}, 0.042  & 30.31/0.122 & 0.024
      
      \multirow{2}{*}{Laptop (S)} & Without noise & \textbf{82.97}  & \textbf{7.34, 10.41} & \textbf{5.16, 9.34} & \textbf{0.040}, \textbf{0.046} & \textbf{0.037}, \textbf{0.042}  & \textbf{30.31}/0.122 & \textbf{0.024}
      \\ \cline{2-9} 
      \specialrule{0em}{1pt}{0pt}
      
      ~ & With noise & {70.04} & {16.01}, {13.27} & {11.47}, {9.52} & {0.082}, {0.067} & {0.075}, {0.065} & {32.84}/\textbf{0.029}   & 0.044
      \\ \cline{1-9} 
      \specialrule{0em}{1pt}{0pt}
      
  \end{tabular}
  }
  \label{tb_exp_abl_oven_noise}
\end{table} 

\begin{figure*}[ht]
  \centering
    \includegraphics[width=1.0\textwidth]{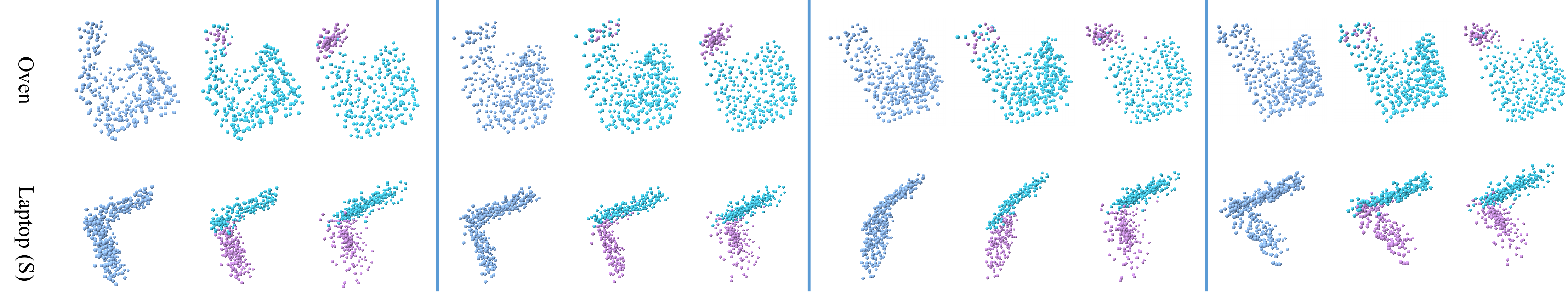}
    \vspace{-5pt}
  \caption{ \footnotesize
    Visualization for the model performance on input data with random noise. Shapes for each three drawn from left to the right are input data corrupted by random normal noise, segmentation, and reconstruction, respectively. 
    We align shapes here just for a better visualization, while they may be put into arbitrary poses for input. 
    % while a random SE(3) rigid transformation will be added to them for input. 
  }
  \label{fig_noose_vis}
  % \vspace{-5pt}
  \vspace{-8pt}
\end{figure*}

Besides testing the performance of the proposed method on partial point clouds with occlusion patterns caused by viewpoint changes, we also test its effectiveness on noisy data. 
Specifically, we add noise for each point in the shape by sampling offsets for its x/y/z coordinates from  normal distributions, \emph{e.g.} $\Delta x \sim \mathcal{N}(0, \sigma^2)$, where we set $\sigma = 0.02$ here. 
Results on Oven and Laptop (S) are presented in Table~\ref{tb_exp_abl_oven_noise}. 
From the table, we can see the degenerated segmentation IoU on Oven's noisy data, while still relatively good part pose estimation performance. 
Another discovery is the even better joint axis orientation prediction, but larger offset prediction perhaps due to the poor segmentation. 
Besides, the shape reconstruction quality also drops a lot, probably due to the randomly shifted point coordinates. 
% For Laptop (S), we can also see the degenerated performance on segmentation IoU and part pose estima
We can observe a similar phenomenon on Laptop (S). 
In Figure~\ref{fig_noose_vis}, we draw some examples for a qualitative understanding w.r.t. model's performance on noise data.

% \todo{normal noise for oven, laptop}

\subsection{Visualization of Part-Level Equivariant Features} \label{sec_appen_part_level}
% \todo{xxx}
% vis for features of points in different parts --- features of 
% vis 
\begin{figure*}[htbp]
  \centering
    \includegraphics[width=0.5\textwidth]{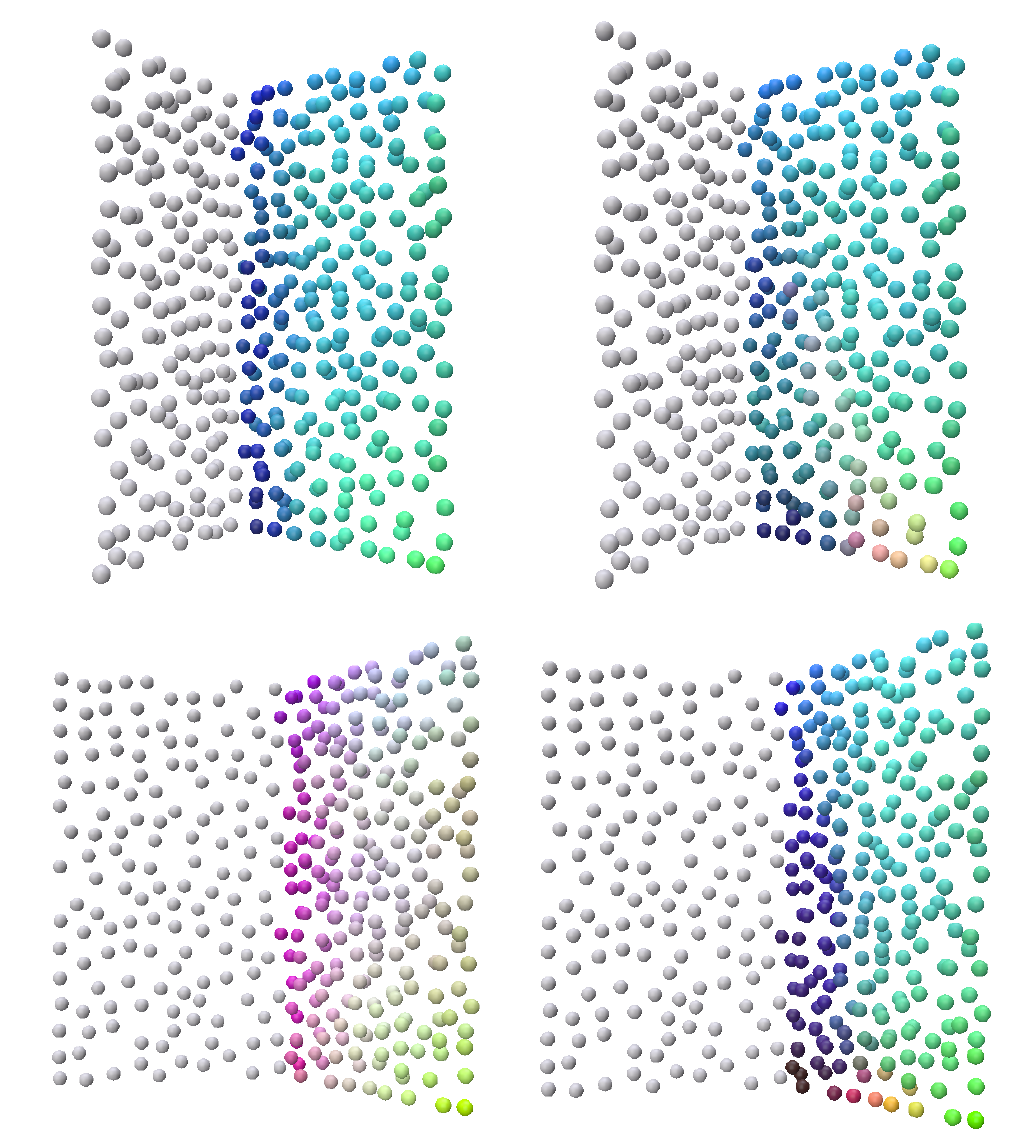}
    \vspace{-5pt}
  \caption{ \footnotesize
    Visualization for an intuitive understanding w.r.t. the difference between the part-level equivariant feature and globa equivariant feature of a specific part. 
    % part features output by the global equivariant network and those output by the part-level equivariant network of the selected mode. 
    Visualized features are obtained by using the PCA algorithm to reduce the feature dimension to 3, 
    % the high-dimensional features to features of dimension 3, 
    which are further normalized to the range of [0, 1]. 
    We only draw point features of the non-motion part with the moving part in gray. 
    Features drawn on the left global equivariant features while those on the right are from the part-level equivariant network. 
    % are those output by the global equivariant network while those on the right are from the designed part-level equivariant network. 
    % We can see that the part-level equivariant feature is more likely to change to the rigid transformation of its own part, seeming to be more suitable to perform part pose and shape disentanglement.
    % Besides, it seems that such features are more suitable to perform part pose and shape disentanglement. 
    % Shapes in the first column are input shapes (we align them for better visualization). 
    % Embedded 2D points drawn in the middle column are dimensional reduced features output by the global equivariant network, while those in the third column are output by part-level equivariant network. 
    % \textcolor{orange}{Orange} 2D points (\textcolor{orange}{trace 1}) drawn in right two columns are embedded features of \textcolor{mypurple}{purple} points in the shape, while \textcolor{myclassicblue}{blue} ones  (\textcolor{myclassicblue}{trace 0}) are for \textcolor{mylakeblue}{deep sky blue} points. 
  }
  \label{fig_part_level_vis_laptop}
  % \vspace{-5pt}
  \vspace{-8pt}
\end{figure*}

Aiming for an intuitive understanding w.r.t. the property output by the designed part-level equivariant network, we draw features output by the global equivariant network and part-level equivariant network for some laptop samples in Figure~\ref{fig_part_level_vis_laptop}. 
From the figure, we can see that the point features of the non-motion part (base) do not change a lot when the moving part (display) rotates an angle. 
That echoes the wish for the part-level equivariance design to disentangle other parts' rigid transformation from the current part's feature learning.
% the difference between features output by the global equivariant network and those by the part-level equivariant network, we draw some samples with different articulation states out in Figure~\ref{fig_part_level_vis_laptop}. 

\subsection{Evaluation Strategy for Category-Level Articulated Object Poses} \label{sec_appen_eval}
To evaluate the category-level part pose estimation performance of our model, we adopt the evaluation strategy used in~\cite{li2021leveraging}. 

For part-based metrics, we first feed a set of train shapes in the canonical articulation states and canonical object pose state to get a set of per-part pose predictions $\{ P_i \}$. 
Then we can calculate the residual pose $\hat{P}_i$  for each part $i$ from the canonical part space defined by human to the canonical part space defined by the network from the pose prediction set (via RANSAC). 
After that, predicted pose from the canonical part space defined by human can be computed by applying the inverse residual pose estimation on the estimated per-part pose, \emph{e.g.} $P_i \leftarrow \hat{P}_i^{-1} P_i$. 
% For joint parameter
When calculating the rotation and translation from part shape $X_1$ to $X_2$, we first centralize their bounding boxes ($\overline{X_1}$ and $\overline{X_2}$, respectively). 
Then, the transformation from $\overline{X_1}$ to $\overline{X_2}$ is taken as the transformation from  $X_1$ to $X_2$. 

For joint parameters, we take the angle error between the predicted joint axis orientation and the ground-truth axis orientation as the metric for joint axis orientation prediction. Metric for joint position prediction is set to the minimum line-to-line distance, following~\cite{li2020category}. 
Only joint axis orientation prediction error is computed for prismatic joints. 
% them by putting their bounding box center to the zero point ($\overline{X_1}$ and $\overline{X_2}$, respectively) 
% and then taking the transformation from $\overline{X_1}$ to $\overline{X_2}$ as their transformation. 

\section{Discussion on Part Symmetry} \label{sec_appen_symmetric_parts}
In this section, we discuss the part-symmetry-related problem that one would encounter in the part pose estimation problem. 
For rigid objects, the pose of a shape is ambiguous for symmetric shapes. 
To say a shape $X$ is symmetric, we mean that there is a non-trivial SE(3) transformation $S_{A_0}$ such that $X = S_{A_0}[X]$. 
In those cases, the performance of the pose estimation algorithm may degenerate due to ambiguous poses. 
It is a reasonable phenomenon, however.  
% For symmetric 
But for articulated objects, we may have symmetric parts even if the whole shape is not a symmetric one. 
For those shapes, we still expect for accurate part pose estimation. 
It indicates that estimating part poses for each part individually is not reasonable due to part pose ambiguity. 
That's why we choose to model the relationship between parts, or specifically, the kinematic chain, joint parameters. 
Without such object-level inter-part modeling, we cannot get accurate part poses by estimating their pose individually, even using ground-truth segmentation. 
The comparison between Oracle EPN and our method in Table~\ref{tb_exp_pose_cmp_other_baseline} can demonstrate this point to some extent. 

% \section{Appendix}

\end{document}